\documentclass[letter, 10pt]{IEEEtran}
\usepackage{graphicx}

\usepackage{subcaption}

\newcommand{\subsubfloat}[2]{%
  \begin{tabular}{@{}c@{}}#1\\#2\end{tabular}%
}

\usepackage{multicol}
\usepackage{booktabs}
\usepackage{amsmath,amssymb}
\usepackage{multirow}
\newcommand{\tabincell}[2]{\begin{tabular}{@{}#1@{}}#2\end{tabular}}

\usepackage[font=small]{caption}
\usepackage{amsmath}
\usepackage{amssymb}
\usepackage{float}

\usepackage{lettrine}
\usepackage{algorithm}
\usepackage{algpseudocode}
\usepackage{cite}
\usepackage{filecontents}
\usepackage{lipsum}
\usepackage{color}
\usepackage{esdiff}
\usepackage{epstopdf}
\usepackage{balance}
\usepackage[utf8]{inputenc}
\usepackage[english]{babel}
\usepackage{amsthm}

\newtheorem{defn}{Definition}

 \newcommand{\Real}{\mathbb{R}}

 \definecolor{pink}{rgb}{1, 0, 1}
 \definecolor{orange}{rgb}{1, 0.7529, 0}
 
 \usepackage{balance}

\begin{document}

\title{

\Huge {SMARRT: Self-Repairing Motion-Reactive Anytime\\ RRT for Dynamic Environments}\\
}

\author{ \begin{tabular}{cccccccccc}
{Zongyuan Shen$^\dag$$^\star$} & {James Wilson$^\dag$} & {Ryan Harvey$^\dag$} & {Shalabh Gupta$^\dag$}
\end{tabular}\vspace{-18pt}
\thanks {$^\dag$Dept. of Electrical and Computer Engineering, University of Connecticut, Storrs, CT 06269, USA.}
\thanks {$^\star$Corresponding author (email: zongyuan.shen@uconn.edu)}
}

\maketitle

\begin{abstract}
This paper addresses the fast replanning problem in dynamic environments with moving obstacles. 
Since for randomly moving obstacles the future states are unpredictable, the proposed method, called SMARRT, reacts to obstacle motions and revises the path in real-time based on the current interfering obstacle state (i.e., position and velocity). SMARRT is fast and efficient and performs collision checking only on the partial path segment close to the robot within a feasibility checking horizon. If the path is infeasible, then tree parts associated with the path inside the horizon are pruned while maintaining the maximal tree structure of already-explored regions. 
Then, a multi-resolution utility map is created to capture the environmental information used to compute the replanning utility for each cell on the multi-scale tiling. 
A hierarchical searching method is applied on the map to find the sampling cell efficiently. 
Finally, uniform samples are drawn within the sampling cell for fast replanning. 
The SMARRT method is validated via simulation runs, and the results are evaluated in comparison to four existing methods. 
The SMARRT method yields significant improvements in travel time, replanning time, and success rate compared against the existing methods. 

\end{abstract}

\begin{IEEEkeywords}
Motion planning, real-time, dynamic environments, replanning.
\end{IEEEkeywords}

\IEEEpeerreviewmaketitle

\thispagestyle{empty}
\section{Introduction}

Motion planning is the fundamental task of determining a collision-free trajectory that drives the robot to a target state in the configuration space while minimizing a user-defined cost function, e.g. trajectory length. 
It has a wide range of applications for autonomous vehicles, such as seabed mapping\cite{shen2020ct,shen2016,shen2017}, structural inspection\cite{song2020online}, autonomous urban driving\cite{kuwata2009real}, mine hunting\cite{MGR11}, and other robotic tasks. Frequent replanning is necessary for the path planner to ensure that the robot successfully completes its mission, as real world applications often involve avoiding unpredictably moving obstacles. It is thus of critical importance to ensure online and rapid replanning of the robot's path in order to guarantee safe and reliable operation as new information becomes available.


Various sampling-based reactive replanning methods have been proposed in literature (see Section~\ref{sec:review}), however, the existing methods have several limitations.
First of all, they check the collision status of either the entire tree or the entire path. This is inefficient in dynamic environments since the current obstacle information could become invalid quickly.
Moreover, the collision checking procedure is known to be the primary computational bottleneck of sampling-based motion planning algorithms\cite{lavalle2006planning}. 
Thus, their performance can be degraded in highly dynamic and complex environments.
Besides, the existing approaches prune or inactivate the tree parts that are not obstructed by the obstacles to maintain a single or double-tree structure.
This results in repeated exploration of the already-explored area in the future, which is wasteful. 
Last but not least, the existing approaches apply a standard uniform sampler and/or sample biasing method to generate the samples needed to either grow the single tree or repair connectivity between disjoint trees. This growth can easily create redundant samples which are not helpful for replanning.



To address the above limitations, this paper presents Self-repairing Motion-reactive Anytime RRT (SMARRT) that can provide rapid replanning in highly dynamic environments.
We introduce an efficient feasibility checking method which only checks the feasibility of path segments close to the robot, while the remaining segments can forgo checking feasibility to reduce computational time.
If the path is infeasible, a feasibility checking horizon with limited size is constructed around the robot.
Then, only the tree parts inside the horizon are checked against collision, and colliding parts are pruned while maintaining the maximal tree structure of already-explored regions.
After the removal of colliding parts, the \textit{floodfill} algorithm \cite{Heckbert1990} is applied to identify the set of disjoint trees, meanwhile, a multi-resolution utility map is constructed to capture essential information of the environment, and computes the utility for each cell on the multi-scale tiling.
Then, a sampling cell is searched in a bottom-up manner on the map; uniform samples are generated randomly within this cell to join to all of the disjoint trees by connecting them to the nearest nodes in this cell for fast replanning.
Moreover, upon reaching the waypoint, the robot switches to a better path found in the local neighborhood if possible. Fig.~\ref{fig:example} shows an illustrative example of SMARRT algorithm in a dynamic environment. 



The main contribution is summarized as follows.

\begin{itemize}
    \item Efficient feasibility checking and risk-based tree pruning methods that delay the complex computation until it is imperative.
    \item A multi-resolution utility map which captures essential information within the current configuration space to facilitate safe, optimized, and efficient replanning.
    \item A hierarchical sampling cell search method which significantly reduces the computational complexity for real-time applications.
    \item Comprehensive evaluation and comparison with alternative methods, showing significant improvements in terms of success rate, travel time, and replanning time.
\end{itemize}



\begin{table*}[ht!]{}
\caption {Comparison of Key Features of SMARRT with other Algorithms}\label{tab:feature}
\centering
 \begin{tabular}{l l l l l l l l} 
 \toprule
 
\specialrule{0em}{2pt}{2pt}\vspace{4pt}
& SMARRT 
& ERRT\cite{bruce2002real}  
& DRRT\cite{ferguson2006replanning} 
& MP-RRT\cite{zucker2007multipartite} 
& EBGRRT\cite{yuan2020efficient} 
& RRT$^\text{X}$\cite{otte2016rrtx}
& HLRRT*\cite{chen2019horizon} \\ 
\hline 

\specialrule{0em}{3pt}{3pt}
\tabincell{l}{Approach} 
& \tabincell{l}{{Maintains} \\ {disjoint trees} \\ {and searches} \\ {promising area} \\ {for sampling on} \\ {multi-resolution} \\ {utility map for} \\ {fast replanning}} 
& \tabincell{l}{{Grows a new} \\ {tree and uses} \\ {sample biasing}} 
& \tabincell{l}{{Keeps goal-} \\ {rooted subtree} \\ {and uses} \\ {sample biasing}} 
& \tabincell{l}{{Maintains} \\ {disjoint trees} \\ {and actively} \\ {reconnects} \\ {disjoint trees} \\ {to root tree}} 
& \tabincell{l}{{Maintains} \\ {double trees} \\  {and reconnects} \\ {robot's sitting} \\ {tree to root tree} \\ {using heuristics}} 
& \tabincell{l}{{Repairs graph} \\ {by rewiring} \\ {cascade and} \\ {extracts a} \\ {shortest-path-to-} \\ {goal subtree}}
& \tabincell{l}{{Keeps robot-} \\ {rooted subtree} \\ {and uses} \\ {sample biasing}} \\ 

\specialrule{0em}{3pt}{3pt}
\tabincell{l}{{Tree Root}} 
& \tabincell{l}{Goal}
& \tabincell{l}{Robot position} 
& \tabincell{l}{Goal} 
& \tabincell{l}{Robot position} 
& \tabincell{l}{Robot position}
& \tabincell{l}{Goal}
& \tabincell{l}{Robot position}\\

\specialrule{0em}{3pt}{3pt}
\tabincell{l}{{Feasibility} \\ {Checking} \\ {for existing} \\ {tree or path}} 
& \tabincell{l}{{Checks path} \\ {segments nearby} \\ {and then tree} \\ {parts nearby if} \\ {path is infeasible}}
& \tabincell{l}{Entire path} 
& \tabincell{l}{Entire tree} 
& \tabincell{l}{Entire tree} 
& \tabincell{l}{Entire path} 
& \tabincell{l}{Entire graph} 
& \tabincell{l}{Entire tree} \\

\specialrule{0em}{3pt}{3pt}
\tabincell{l}{{Tree} \\ {Pruning}} 
& \tabincell{l}{{Prunes colliding} \\ {nodes}}
& \tabincell{l}{{Discards entire} \\ {tree}} 
& \tabincell{l}{{Prunes colliding} \\ {nodes and all} \\ {successors}} 
& \tabincell{l}{{Prunes colliding} \\ {nodes}}
& \tabincell{l}{{Prunes colliding} \\ {nodes and some} \\ {successors that} \\ {are unreachable} \\ {from the goal}}
& \tabincell{l}{{Inactivates} \\ {colliding nodes}}
& \tabincell{l}{{Prunes colliding} \\ {nodes and all} \\ {successors}} \\

\specialrule{0em}{3pt}{3pt} \vspace{3pt}
\tabincell{l}{{Sampling} \\ {Strategy}} 
& \tabincell{l}{{Creates random} \\{samples within} \\ {promising region}}
& \tabincell{l}{{Selects samples} \\ {from previous} \\ {path, goal and} \\ {random point}} 
& \tabincell{l}{{Selects samples} \\ {from previous} \\ {path, goal and} \\ {random point}} 
& \tabincell{l}{{Selects samples} \\ {from disjoint} \\ {tree roots, goal} \\ {and random point}} 
& \tabincell{l}{{Selects samples} \\ {from previous} \\ {path, goal and} \\ {random point}} 
& \tabincell{l}{{Standard} \\ {sampler}}
& \tabincell{l}{{Selects samples} \\ {from a learned} \\ {distribution, goal} \\ {and random point}}\\


 \toprule
 \end{tabular}
 \vspace{-10pt}
 \end{table*}

The remaining work is organized as follows: Section~\ref{sec:review} reviews the existing work on sampling-based replanning algorithms; Section~\ref{sec:problem} formulates the problem, while Section~\ref{sec:solution} describes the proposed method; Section~\ref{sec:results} presents the results and Section~\ref{sec:conclusions} provides the concluding remarks.

\section{Related Works}
\label{sec:review}



Sampling-based methods generate samples randomly in the configuration space, construct a graph structure to capture connectivity between different configurations, and search a solution on it \cite{elbanhawi2014sampling}. Specifically, they are very useful for online planning in dynamic environments.
Online methods are characterized as active or reactive. 
Active strategies predict the future trajectory of a moving obstacle for a fixed time duration and generate a collision-free trajectory for the robot; however, the performance can degrade if the predicted information is incomplete or incorrect. 
In contrast, reactive strategies plan the path based only on the obstacle information at the current time, and replan the path whenever the obstacle information changes. 
Some sampling-based reactive approaches are discussed below.



Bruce and Veloso\cite{bruce2002real} presented Extended RRT (ERRT), which regrows a new tree from scratch whenever the current path is infeasible.
Ferguson et al.\cite{ferguson2006replanning} proposed Dynamic RRT (DRRT), which prunes the colliding nodes and their successors from the tree and grows the tree using RRT algorithm. 
Zucker et al.\cite{zucker2007multipartite} presented Multipartite RRT (MP-RRT), which only prunes the colliding nodes, resulting in multiple disjoint trees. 
Then, this method grows the main tree and attempts to reconnect other disjoint trees to the main tree at each iteration.
Yuan et al.\cite{yuan2020efficient} proposed Efficient Bias-goal Factor RRT (EBG-RRT), which checks the collision status for the current path, prunes the colliding nodes and their successors, while maintaining the usable part of the solution path connected to the goal; resulting in a double tree structure. Then, several heuristics are used to for tree reconnection.



Recently, several variants of the RRT$^\star$ have been presented for the reactive replanning problem.
Otte et al. \cite{otte2016rrtx} proposed RRT$^\text{X}$, an algorithm which utilizes a rapidly-exploring random graph (RRG)\cite{karaman2011sampling} to explore the search area. 
It maintains a well-defined and connected structure over the entire explored region; however, re-optimizing the connections in the entire graph when the environment changes, makes it inefficient in highly dynamic settings.
Chen et al.\cite{chen2019horizon} presented the Horizon-based Lazy RRT$^\star$ (HL-RRT$^\star$) algorithm. In this method, when parts of the search tree are made invalid due to moving obstacles, they are pruned and new samples are drawn to find a new path \textit{via} a trained Gaussian mixture model.
While this can find a new solution quickly, the machine learning based sampler cannot guarantee consistent performance.
Furthermore, this method removes the invalid parts, which are then disconnected from the main tree, to maintain a single tree structure; resulting in repeated exploration of the same area.  

    
         

Overall, these approaches have several limitations: 1) checking the feasibility of either the entire tree, (DRRT, RRT$^\text{X}$, and HLRRT*) or the entire path, (ERRT and EBGRRT), is inefficient in the dynamic environments, since the current obstacle information could become invalid soon after; 2) pruning or inactivating the parts which are not in collisions with the obstacles to maintain single tree, (ERRT, DRRT, RRT$^\text{X}$, and HLRRT*) or double tree structure (EBGRRT), which results in repeated exploration of the already-explored area; and 3) applying uniform sampler and/or sample biasing method for replanning, which creates redundant samples that are not helpful to reconnect the trees.
In contrast, we proposed a feasible sampling-based motion replanning method which only checks the feasibility of the tree parts close to the robot.  This method removes the colliding nodes while retaining the maximal tree structure for reconnection and expansion simultaneously, and also facilitates sampling inside promising regions for fast replanning. 
First, we presents the problem statement as follows. 

\section{Problem Statement}
\label{sec:problem}

Let $\mathcal{X}$ be the configuration space of the robot. Let $\mathcal{X}_{obs} \subset \mathcal{X}$ be the obstacle space, such that $\mathcal{X}_{free} = \mathcal{X} \slash \mathcal{X}_{obs}$ is the free space. The location of the robot at time $t$ is denoted by $x_{robot}(t) \in \mathcal{X}_{free}$. First, we define the trajectory as follows.

\begin{defn}[Trajectory]\label{define:trajectory}
Given initial condition $x_{init}$ and the final condition $x_{goal}$, a continuous function $\sigma:[t_0,t_f] \rightarrow \mathcal{X}_{free}$ of bounded variation is called a trajectory if

\begin{itemize}
\item $\sigma(t) \in \mathcal{X}_{free}, t \in [t_0,t_f]$.
\item $\sigma(t_0) = x_{init}, \sigma(t_f) = x_{goal}$.
\item $\sigma$ satisfies the kinodynamic constraints of the robot. In this paper, we only consider the holonomic vehicle.
\end{itemize}

\end{defn}

    
         

During the online operation, the obstacle space changes as a function of time and/or robot location, i.e., $\Delta\mathcal{X}_{obs} = f(t,x_{robot})$. Then, the dynamic environment is defined below.

\begin{defn}[Dynamic Environment]\label{define:dynamic}
An environment is said to be dynamic if it has an unpredictably changing obstacle space, i.e., the function $f$ is \textit{a priori} unknown. Based on the definition, dynamic environments can contain obstacles that unpredictably appear, disappear, or move.
\end{defn}

In this paper, we only consider randomly moving obstacles whose behavior is unpredictable. Then, we define the feasible replanning problem as follows.

\begin{defn}[Feasible Replanning Problem]\label{define:trajectory}
Given $\mathcal{X}_{obs}$, $\mathcal{X}_{free}$, $x_{goal}$, $x_{robot}(t_{0})=x_{init}$, and unknown $\Delta\mathcal{X}_{obs} = f(t,x_{robot})$, find the feasible trajectory  $\sigma:[t_{cur},t_f] \rightarrow \mathcal{X}_{free}$ with robot's current position $x_{robot}(t_{cur})$ as its start, until $x_{robot}(t_{cur}) = x_{goal}$, constantly update $x_{robot}(t_{cur})$ along $\sigma$ while recalculating $\sigma$ whenever $\Delta \mathcal{X}_{obs} \neq \emptyset$. 


\end{defn}

\section{SMARRT Algorithm}
\label{sec:solution}

The key ideas behind SMARRT are 1) the utilization of an efficient path feasibility checker within the local neighborhood of the robot, and 2) smart and rapid replanning if the current path is (likely to become) infeasible due to a nearby moving obstacle. This is achieved via the following steps. First, in order to facilitate rapid replanning, we initialize a \textit{multi-resolution utility map} that provides an efficient spatial indexing of the states in the search tree. Next, an initial search tree is created using the RRT path planner considering only the static obstacles. As sampled states are added to the search tree, their locations are simultaneously indexed in the multi-resolution map. Then, the robot moves along the found path towards the goal while constantly checking its local feasibility against dynamic obstacles using a \textit{feasibility checking horizon.} If the robot path is likely to become obstructed in the near-future, the infeasible states are pruned. This results in a set of disjoint trees that need to be repaired so that a new path can be found. Using the multi-resolution map, regions that contain states from multiple disjoint trees are identified since these regions are the most likely candidates for rapid repair. Then, the utility of these regions are computed based on the estimated path length from the robot's current position to the goal while passing through this region. Finally, samples are placed in the regions with the highest utility to repair the tree and guide the robot towards its objective. Details of the SMARRT algorithm are provided below.

\subsection{Initialization of the Multi-Resolution Utility Map}
The multi-resolution map is a hierarchical tiled representation of the workspace. Let $\mathcal{A}\subset \mathbb{R}^2$ be the workspace. A tiling on $\mathcal{A}$ is defined as follows:

\begin{defn}[Tiling]\label{define:Tiling}
A set $\mathcal{T} = \{\tau_{\gamma} \subset \mathbb{R}^2:{\gamma}= 1,\ldots|\mathcal{T}|\}$, is a tiling of $\mathcal{A}$, if its elements, called tiles (or cells), have mutually exclusive interiors and cover $\mathcal{A}$, i.e., 
\begin{eqnarray*}
& \bullet & \ I\big(\tau_{\gamma}\big) \bigcap I\big(\tau_{\gamma'}\big) =\emptyset, \forall {\gamma},{\gamma'} \in \{1,\ldots|\mathcal{T}|\}, {\gamma} \neq {\gamma'}\\
& \bullet & \ \bigcup_{\gamma=1}^{|\mathcal{T}|}\tau_{\gamma} =\mathcal{A},
\end{eqnarray*}
where $I\big(\tau_{\gamma}\big)$ denotes the interior of the cell $\tau_{\gamma} \in \mathcal{T}$. 
\end{defn}

\begin{figure*}[ht!]
    \centering
    \subfloat{
        \includegraphics[width=0.4\textwidth]{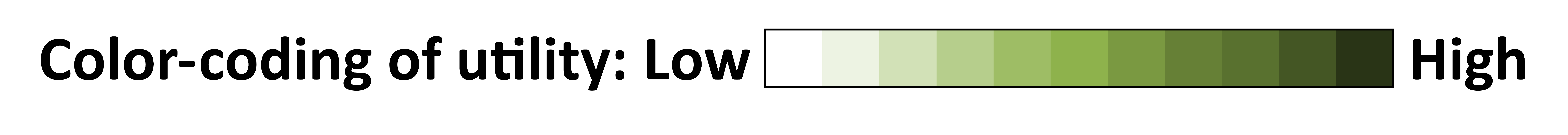}\label{fig:colorbar}}\hfill\vspace{0pt}\\
        \setcounter{subfigure}{0}
    \centering
    \subfloat[Two disjoint trees are created.]{
        \includegraphics[width=0.24\textwidth]{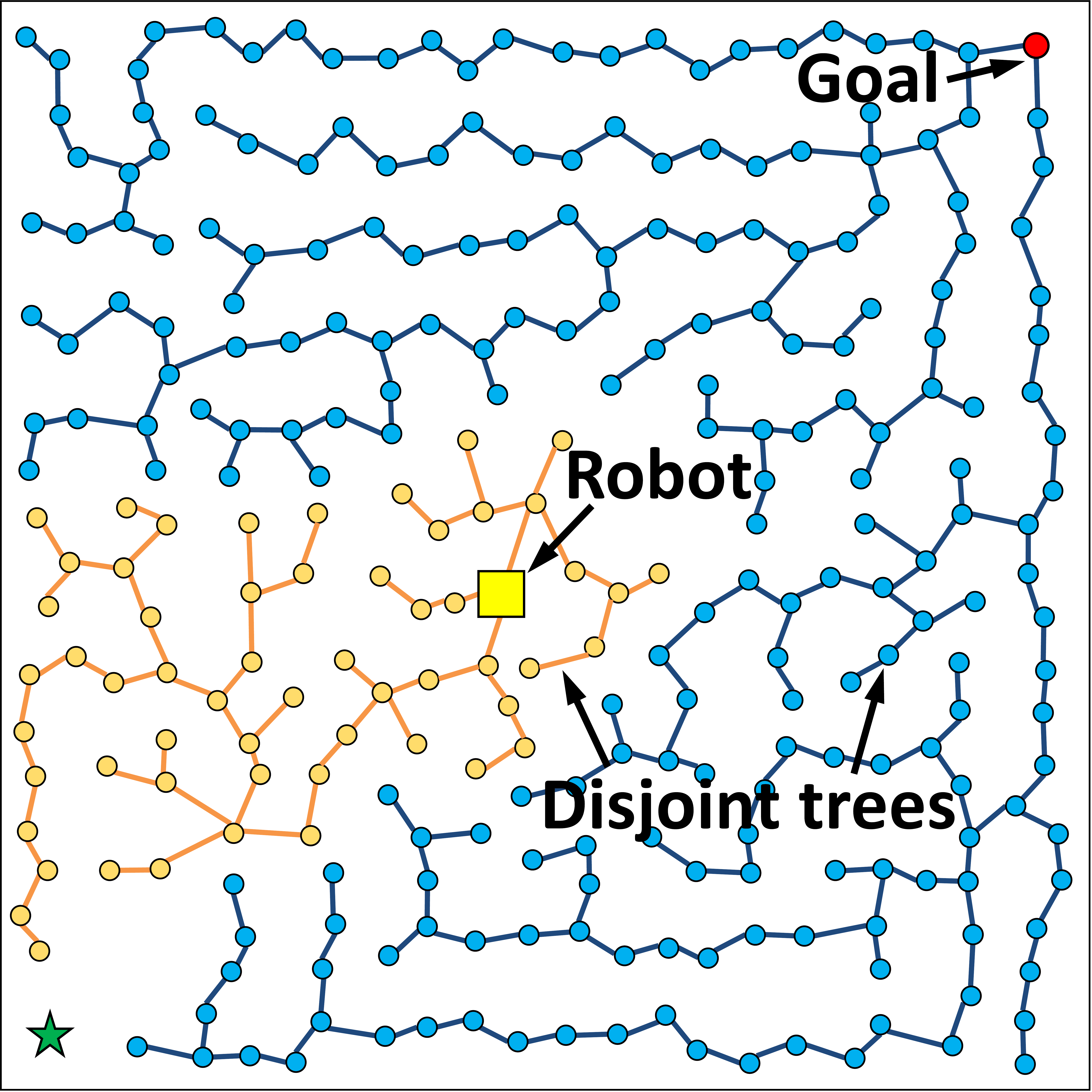}\label{fig:map_part1}}\vspace{0pt}\hspace{-8pt}\quad
         \centering
    \subfloat[Utility map at level 0.]{
         \includegraphics[width=0.24\textwidth]{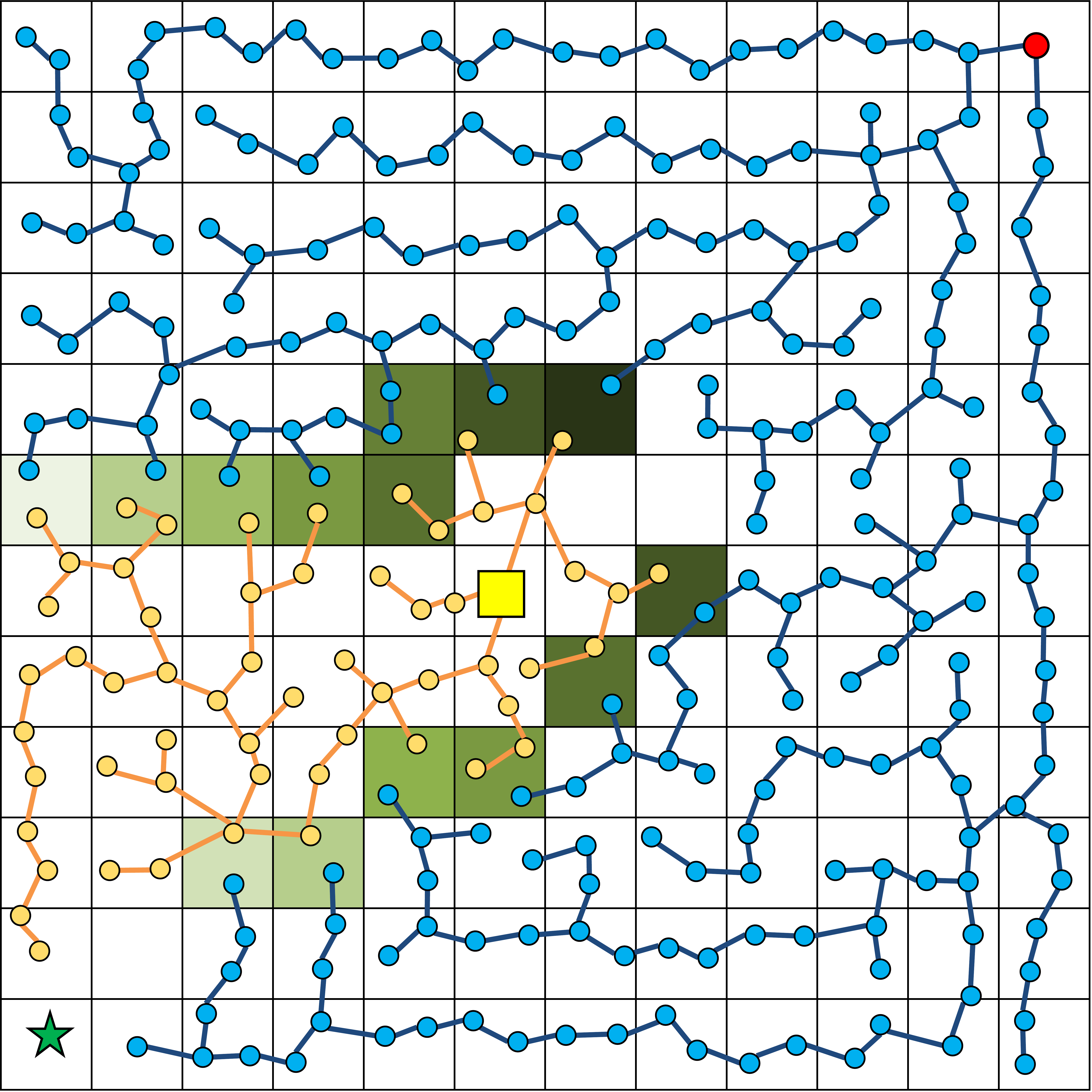}\label{fig:map_part2}}\hspace{-8pt}\quad
         \centering
    \subfloat[Utility map at level 1.]{
         \includegraphics[width=0.24\textwidth]{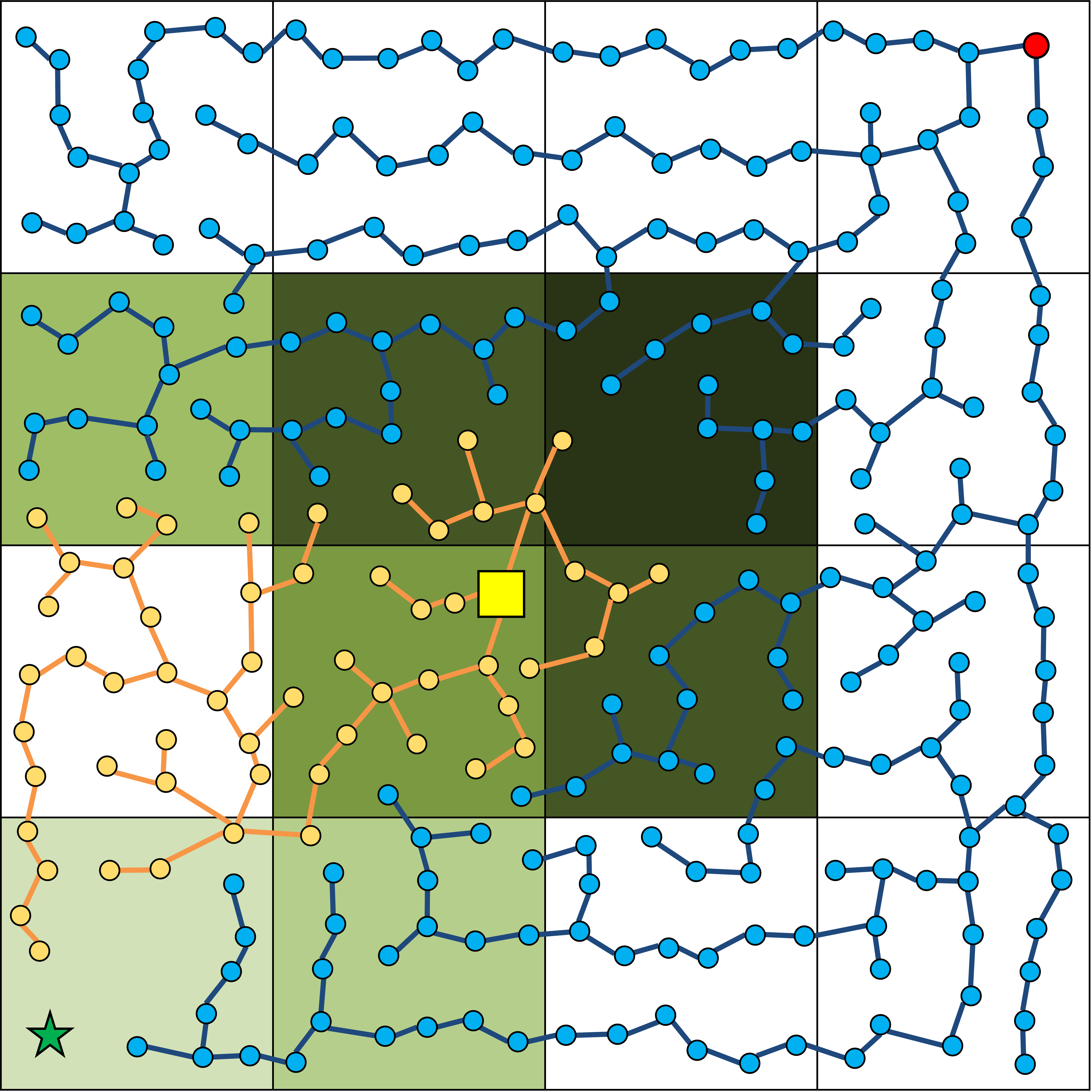}\label{fig:map_part3}}\hspace{-8pt}\quad
         \centering
    \subfloat[Utility map at level 2.]{
        \includegraphics[width=0.24\textwidth]{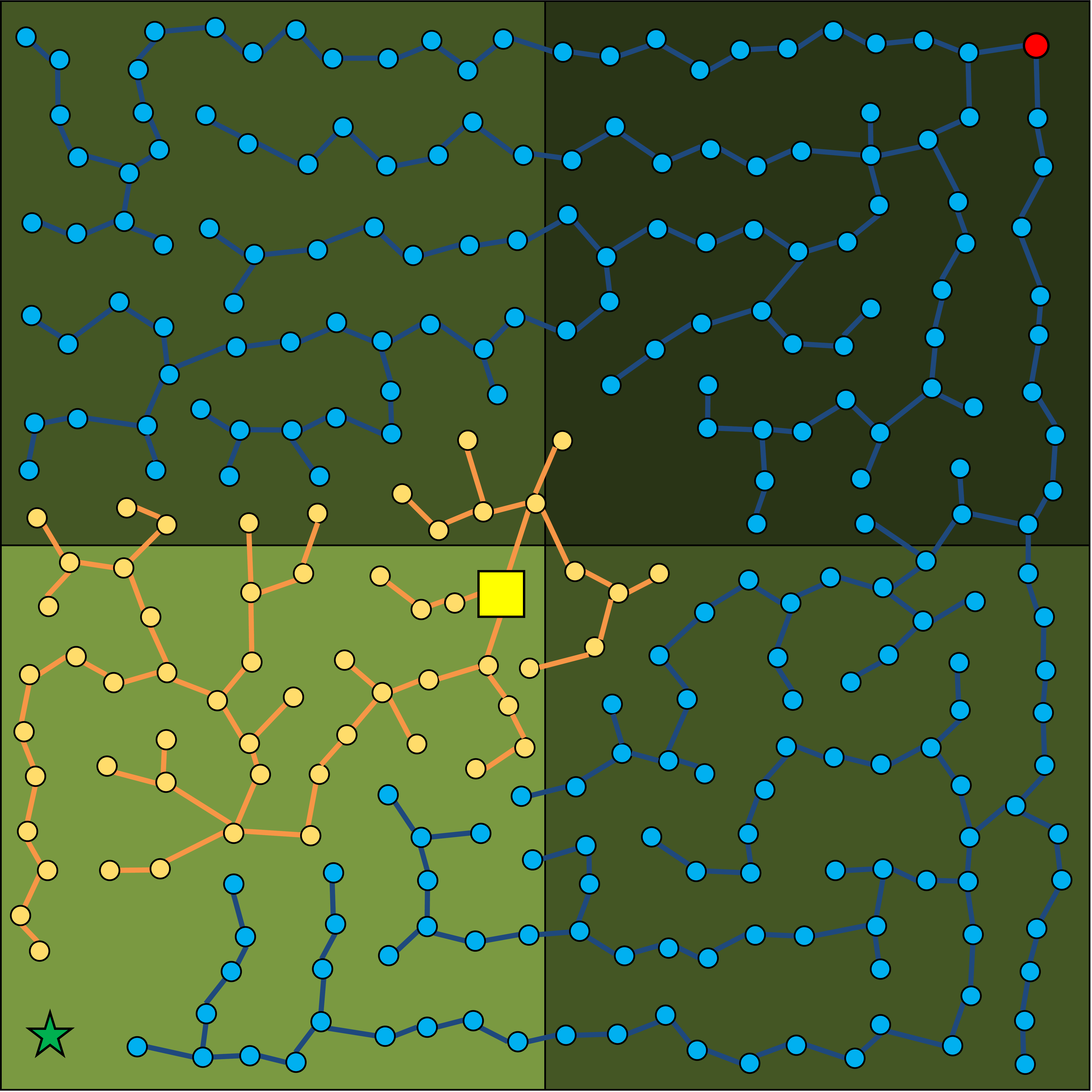}\label{fig:map_part4}}\\ 
        
          \caption{An example of multi-resolution utility map.}\label{fig:map}
 \end{figure*}

The search area $\mathcal{A}$ is decomposed recursively to generate a hierarchical multi-scale tiling (MST) structure, as shown in Fig.~\ref{fig:map}. The purpose of the MST is to enable incremental utilization of the knowledge of the environment map at different scales as needed. The MST provides two basic advantages:

\begin{itemize}
    \item Reduction of computational complexity and
    \item Determination of the replanning area quickly,
\end{itemize}
which will be discussed in details later in Section~\ref{sampling_cell}.

In order to create the MST, the finest level of the MST, denoted as $\mathcal{T}^0$, is first constructed according to a user-defined minimum cell size, as seen in Fig.~\ref{fig:map_part2}. Without loss of generality, the cell size is defined such that there is an even number of cells, denoted as $N_x$ and $N_y$, along the $x$ and $y$-axes, respectively. Then, the coarsest level of the MST, denoted as $\mathcal{T}^L$, is constructed by partitioning the search area $\mathcal{A}$ into four equally-sized quadrants such that each quadrant has dimensions $\frac{N_x}{2}\times \frac{N_y}{2}$, as seen in Fig.~\ref{fig:map_part4}. The intermediate levels $\mathcal{T}^1\dots\mathcal{T}^{L-1}$ are constructed as follows. Level $\mathcal{T}^{L-1}$ is created by subdividing the four course cells in $\mathcal{T}^L$ such that each cell has a dimensions $\frac{N_x}{4}\times\frac{N_y}{4}$, giving a total of 16 cells in $\mathcal{T}^{L-1}$. This process is repeated until level $\mathcal{T}^1$ is reached, which has the smallest tiling representation such that its cells are larger than those in $\mathcal{T}^0$, i.e., $\frac{N_x}{2^L}\times\frac{N_y}{2^L} \geq 2\times 2$. Fig.~\ref{fig:map_part3} shows an example of an intermediate tiling level. This procedure generates a MST with levels $\mathcal{T}^0$,$\mathcal{T}^1$, ...$\mathcal{T}^L$ where $\mathcal{T}^{\ell} = \left \{{\tau_{{\gamma}^{\ell}}: {\gamma}^{\ell} = 1,...|\mathcal{T}^{\ell}|}  \right \}, \forall{\ell} \in \left \{0,...L  \right \}$.

After the MST is created, the search tree is created using the RRT path planner. As sampled vehicle states are generated and added to the tree, they are simultaneously indexed by the cells in $\mathcal{T}^0$ that contains these states. Specifically, the set of states that are indexed by cell $\tau_{\gamma^0}$ is defined as $P_{\gamma^0}=\{p_j | p_j\in\tau_{\gamma^0}\}$, where $p_j\in\Real^2$ is location of the vehicle state in the search tree. This is done to enable the quick identification of the key regions for sampling during the replanning phase. 

\subsection{Efficient Feasibility Checking and Risk-based Tree Pruning}

After the initial path to the goal is found, as shown in Fig.~\ref{fig:example_part1}, the robot starts moving along this path and checking for moving obstacles that might collide with the robot. In order to facilitate real-time path replanning to avoid these risky obstacles, the robot must first identify such risks efficiently. It is well-known that collision checking is the primary computational bottleneck of sampling-based motion planning algorithms\cite{lavalle2006planning}.
Moreover, in dynamic environments, the current obstacle information will soon be invalid as the obstacles move.

\begin{figure}[t]
    \centering
    \includegraphics[width=1\columnwidth]{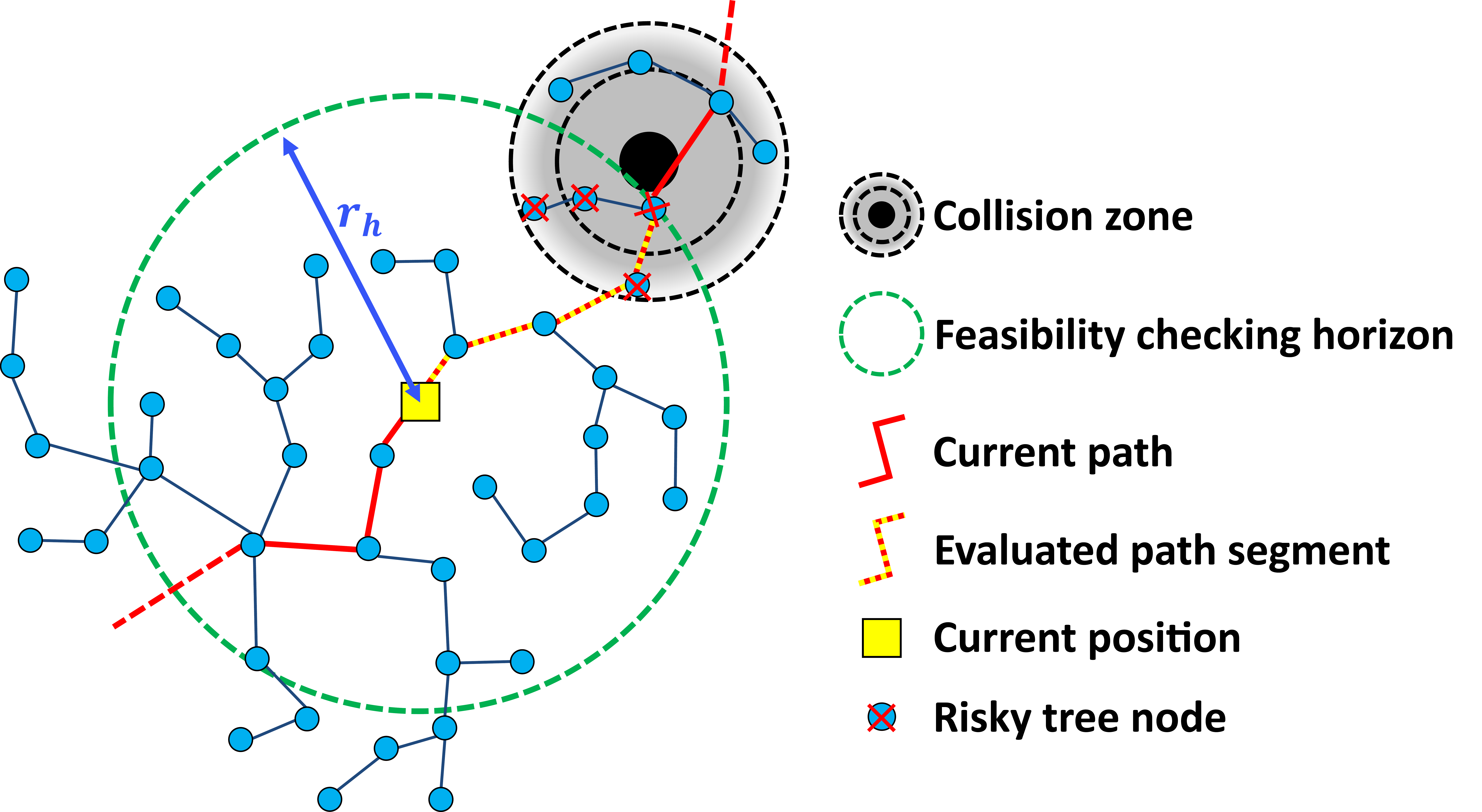}
    \caption{Efficient feasibility checking and risk-based tree pruning.}
   \label{fig:horizon} 
\end{figure}

In this regard, we develop a \textit{feasibility checking horizon} for efficient collision checking and risk-based tree pruning. An example is shown in Fig.~\ref{fig:horizon}. First, the feasibility checking horizon around the robot is determined based on its speed, average replanning time, and the distance it will travel along the path. Collision checking with moving obstacles is restricted to this region since it 1) is large enough to provide sufficient time for the robot to replan safely, and 2) better utilizes the computational resources by eliminating long-range collision checking as far-away obstacles are not a threat to the robot.

The \textit{collision zone} of a moving obstacle is defined similarly to the feasibility checking horizon, except it is centered around said obstacle and based on its current speed. Any nodes that are inside of both the collision zone and the feasibility checking horizon are considered high-risk; these nodes are hence pruned from the tree. In contrast with existing reactive replanning methods, entire branches that connect to these risky nodes are not pruned; instead, \textit{disjoint trees} are formed. This allows SMARRT to retain a maximal-tree structure which allows for efficient exploitation of all previously-explored spaces to repair the search trees quickly. If one of the nodes is on the future path of the robot, then the robot has a high collision-risk with the obstacle and replanning is required. 

Formally, let $\sigma(t)$ be a point on the robot's path at time $t$. Let $t_u$ be the expected replanning time. Since the robot needs to 1) replan and 2) move away from the obstacle, we define $t_h = 2t_u$ to be the feasibility checking time horizon as this provides the minimum sufficient time to escape. Let $p_c = \sigma(t_c)$ and $p_{f} = \sigma(t_c+t_h)$ be the robot's position at the current time instant $t_c$ and at time horizon $t_c+t_h$, respectively. Define $\mathcal{O}=\left \{o_1,o_2,...,o_{|\mathcal{O}|}  \right \}$ as the set of obstacles. Let $v_i(t_c)$ be the speed of obstacle $o_i$ at the current time instant.
Then a set of collision zones $\mathcal{R}^c=\left \{\mathcal{R}_1^c,\mathcal{R}_2^c,...,\mathcal{R}_{|\mathcal{O}|}^c  \right \}$ are constructed at the centroids of each obstacle with radius $r^c_i = 2t_u \times v_i(t_c)$.

Replanning is only required if the the path segment between $p_c$ and $p_f$ intersect with any of the collision zones in $\mathcal{R}^c$. In this scenario, the feasibility checking horizon $\mathcal{R}^f$ is constructed around the robot's current position $p_c$ with radius $r_h = d(p_c,p_{f})$, where $d(p_c,p_{f})$ indicates the Euclidean distance between $p_c$ and $p_{f}$. Any node $p_j\in (\cap_i\mathcal{R}^c_i) \cap \mathcal{R}^f$ is considered risky and are thus pruned. 

\subsection{Real-Time Replanning around Nearby Dynamic Obstacles}

Due to removal of invalid nodes, the original search tree could be broken into multiple disjoint trees. The key feature of SMARRT is its ability to identify cells that are likely to contain nodes from multiple disjoint trees, which enables fast repairing of the tree and thus real-time replanning. To achieve this, disjoint trees must first be identified. Then, the \textit{utility} of each cell is computed which is used by the replanner to inform where the samples should placed to repair the tree while guiding the robot towards the goal.

\subsubsection{Identification of Disjoint Trees} After pruning the risky nodes, the \textit{floodfill} algorithm \cite{Heckbert1990} is applied to identify the set of disjoint trees. Specifically, starting at some random node, the algorithm recursively visits all reachable nodes that belong to the same tree. The above process repeats until each node has been visited, creating a collection of trees $\mathcal{G}=\{\mathcal{G}^1,\dots,\mathcal{G}^{|\mathcal{G}|}\}$. As each node is visited, the corresponding cell that it belongs to records the tree that it is a member of to facilitate faster indexing by the multi-resolution utility map. Specifically, the set of trees that are indexed by cell $\tau_{\gamma^0}$ is defined as $\mathcal{G}_{\gamma^0}=\{\mathcal{G}^g \; | \; \exists p_j\in P_{\gamma^0} \text{ s.t. } p_j\in\mathcal{G}^g , \; g=1,\dots,|\mathcal{G}|\}$. An example of identified disjoint trees after risky node pruning is shown in Fig.~\ref{fig:map_part1}.

\subsubsection{Computation of Cell Utility}
The utility of each cell is computed in a bottom-up manner in the multi-resolution map. Starting at level $\ell=0$, the utility map $\mathcal{U}^0$ is constructed using a two-step procedure. For each cell $\tau_{{\gamma}^0}$ and the corresponding disjoint tree information $\mathcal{G}_{\gamma^0}$, a symbolic state is assigned from the alphabet set $S = \left \{I, V  \right \}$, where $I \equiv invalid$ and $V \equiv valid$. Specifically, the if the cell i) contains nodes from different disjoint trees or ii) has adjacent neighbors that contains nodes from different disjoint trees, it is marked as \textit{V} since creating samples inside these cells will repair the tree instantaneously. Otherwise, the cell is marked as \textit{I}. Then, the utility of cell $\tau_{{\gamma}^0}$ is computed as follows:

\begin{equation}
    U(\tau_{\gamma^0})=\begin{cases}
0 & \text{ if } s(\tau_{\gamma^0})=I\\ 
\frac{1}{d(p_{c},p_{\gamma^0}))+d(p_{\gamma^0},p_g))} &  \text{ if } s(\tau_{\gamma^0})=V
\end{cases}
\end{equation}
where $p_{c}$ and $p_{g}$ indicates the robot's current position and goal position,  respectively, $p_{\gamma^0}$ is the location of the center of cell $\tau_{\gamma^0}$, $s(\tau_{\gamma^0})$ is the symbolic state indicator function, and $d(\cdot,\cdot)$ is the Euclidean distance between two points. If there are no valid cells at level zero, this procedure is repeated for the next level up until at least one valid cell is found. The remaining coarse utility maps are constructed by assigning $\tau_{{\gamma}^{\ell}}$ the maximum utility of the finer cell within $\tau_{{\gamma}^{\ell}}$, i.e.,
\begin{equation}
    U(\tau_{\gamma^{\ell}})=\max_{\tau_{\gamma^{\ell-1}}\in\tau_{\gamma^{\ell}}}U(\tau_{\gamma^{\ell-1}})
\end{equation}
An example of the utility costs are shown in Figs.~\ref{fig:map_part2}-\ref{fig:map_part4}

\begin{figure*}[ht!]
    \subfloat[Initial path is obtained.]{
        \includegraphics[width=0.24\textwidth]{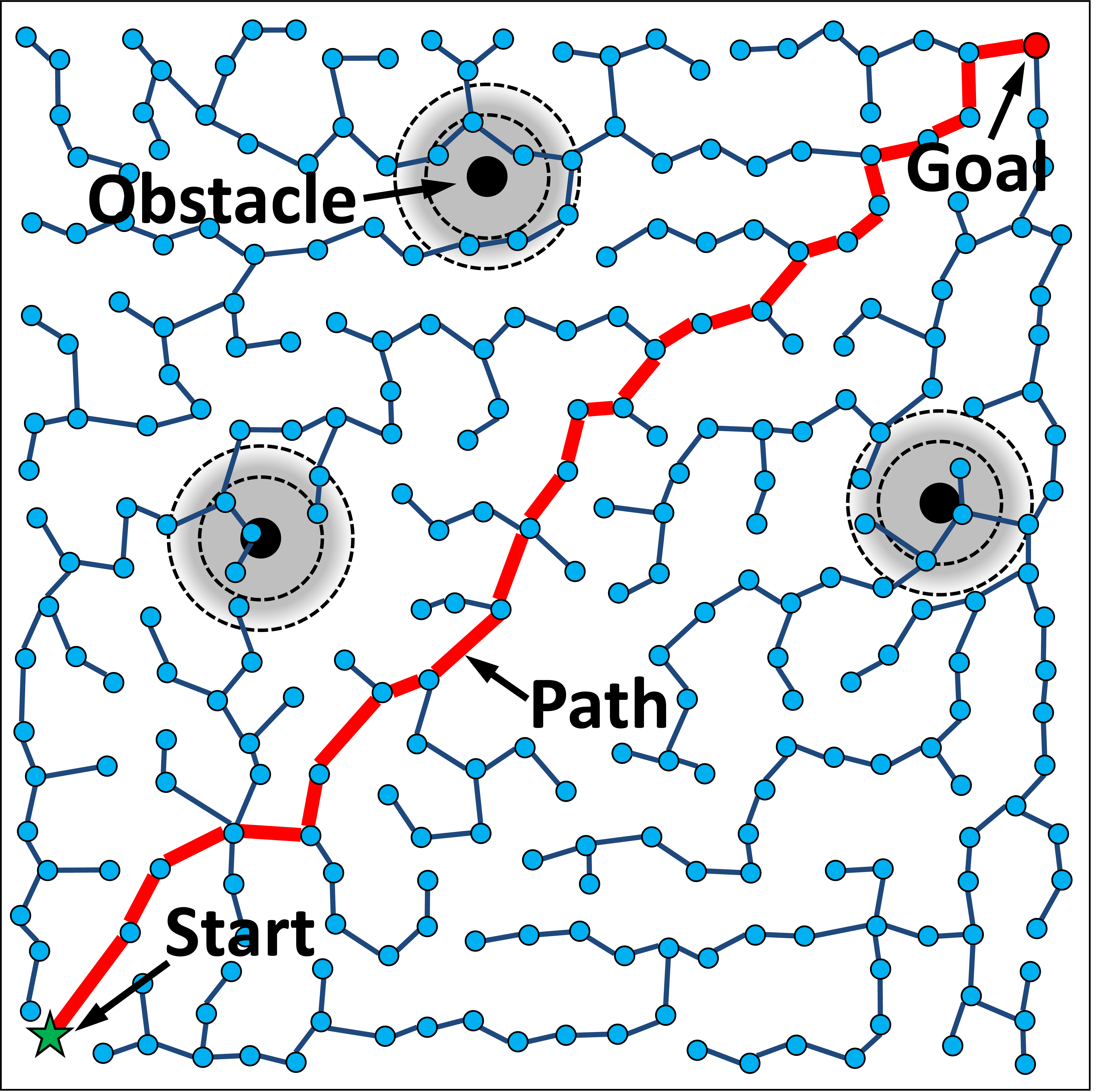}\label{fig:example_part1}}\vspace{0pt}\hspace{-8pt}\quad
         \centering
    \subfloat[Risky nodes are pruned when an obstacle hits the original path.]{
         \includegraphics[width=0.24\textwidth]{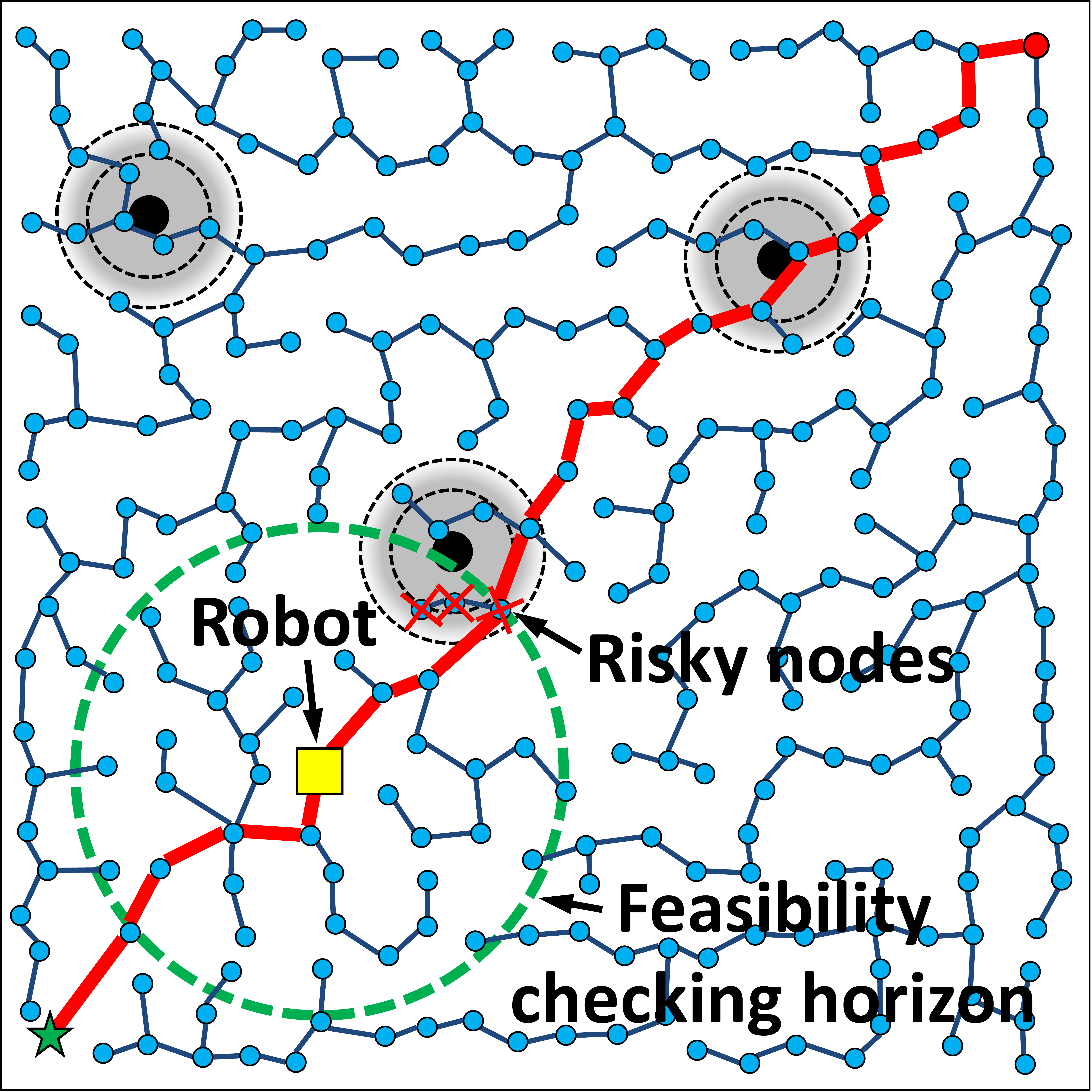}\label{fig:example_part2}}\hspace{-8pt}\quad
         \centering
    \subfloat[Two disjoint trees are created due to risky node pruning.]{
         \includegraphics[width=0.24\textwidth]{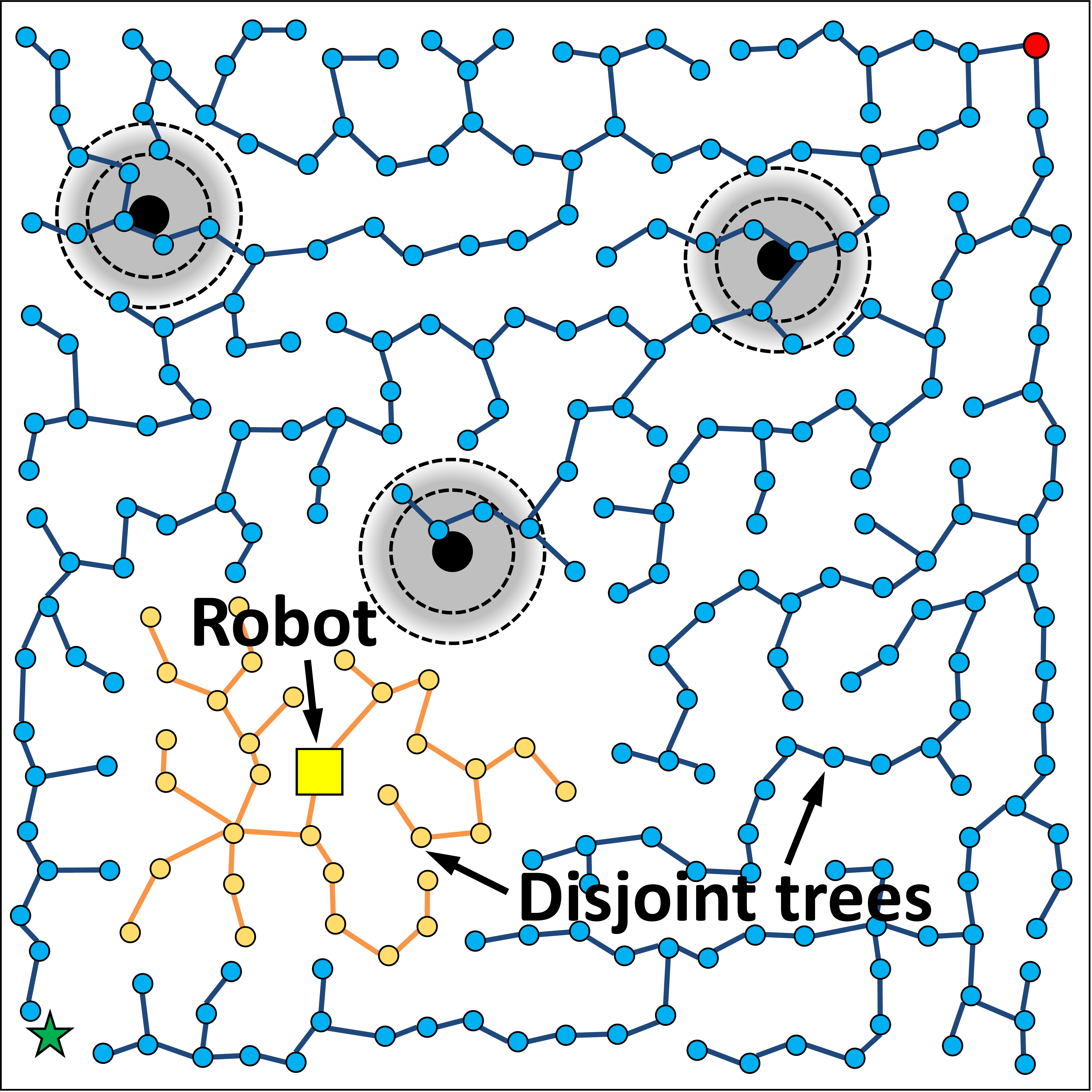}\label{fig:example_part3}}\hspace{-8pt}\quad
         \centering
    \subfloat[A potential tree repairing region is searched in the local neighborhood at level 0 of the tiling. None is found.]{
        \includegraphics[width=0.24\textwidth]{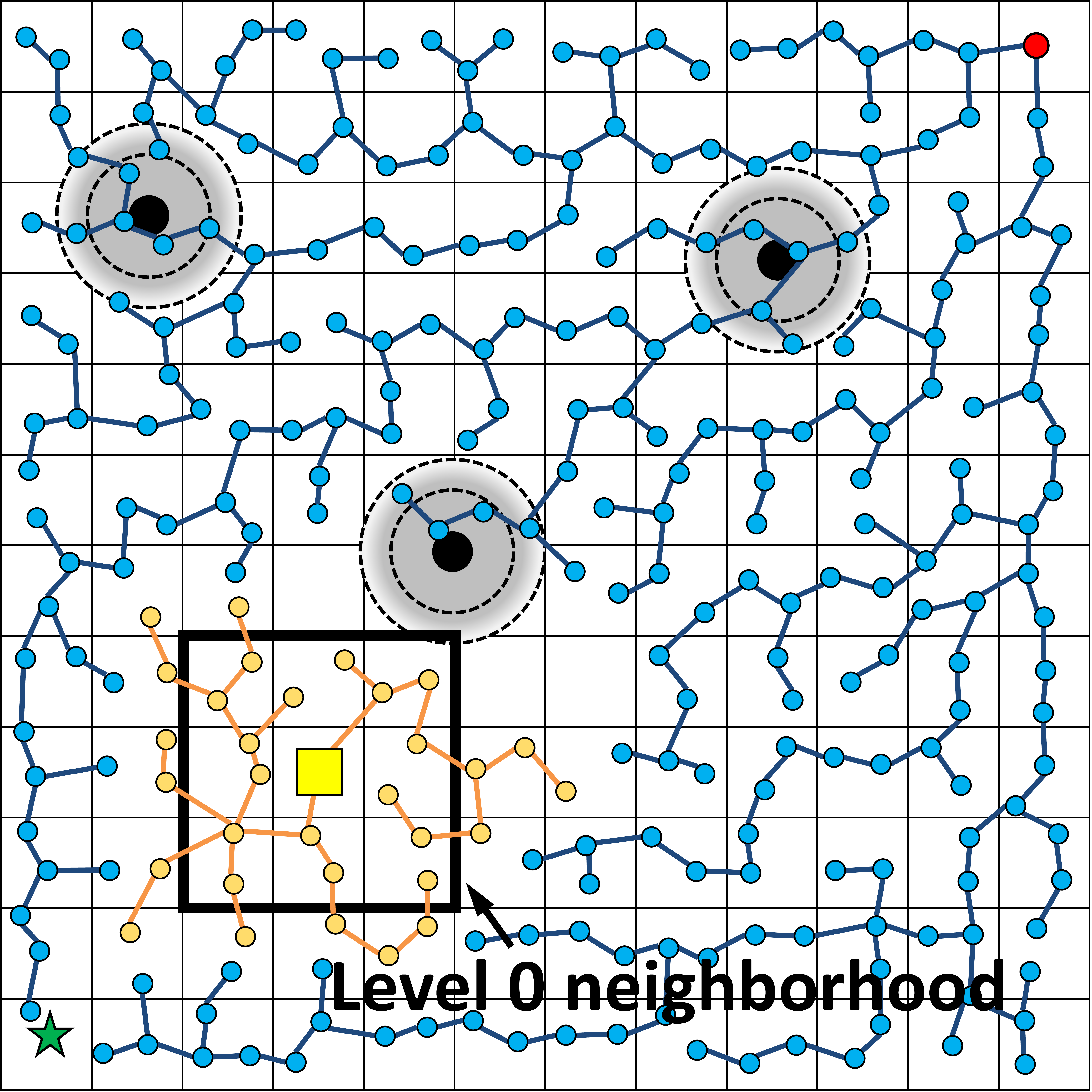}\label{fig:example_part4}}\\ 
        
        \subfloat[Search is expanded to the local neighborhood at level 1 of the tiling. Four potential regions are found and the one with the highest utility is selected.]{
        \includegraphics[width=0.24\textwidth]{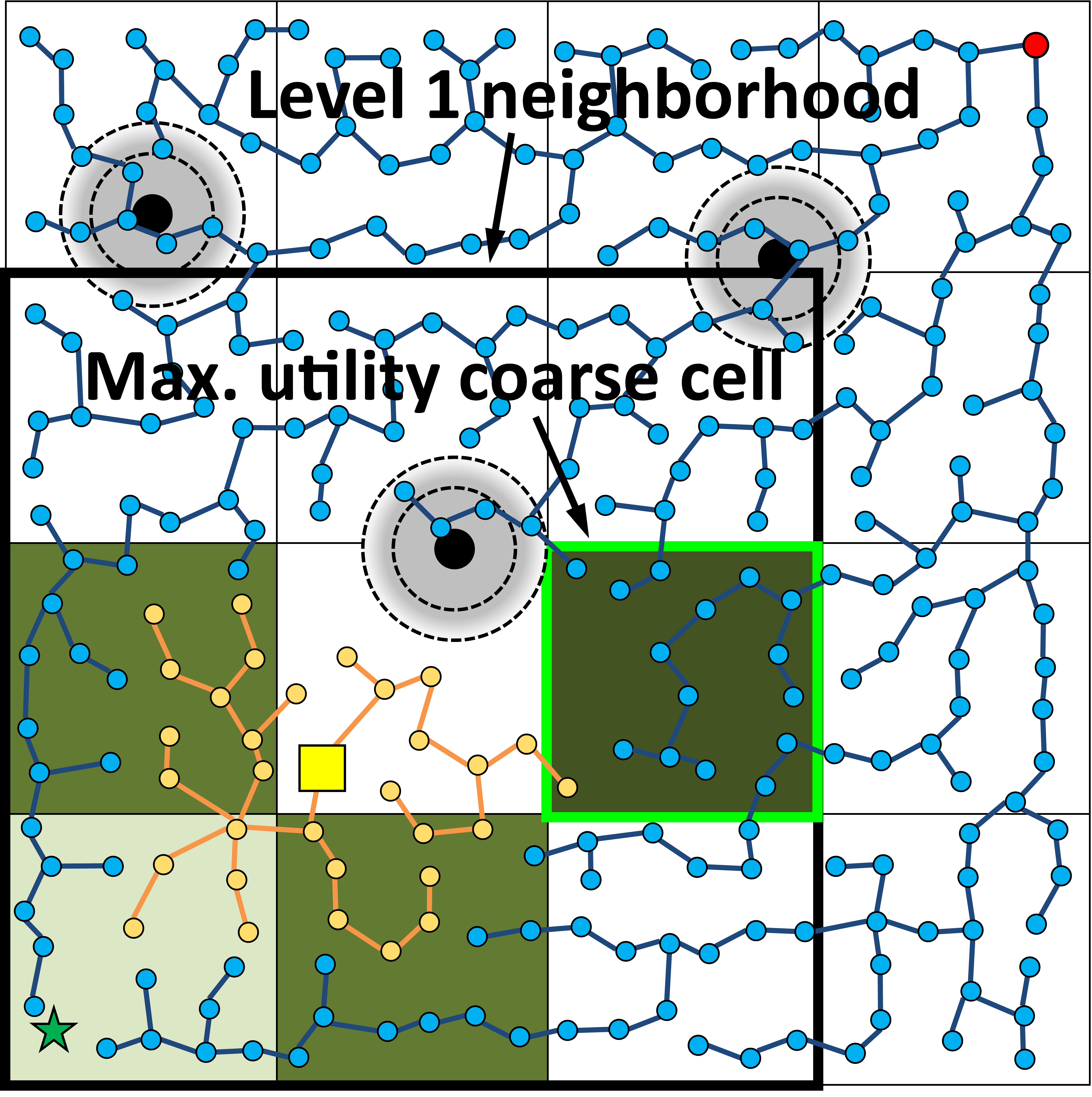}\label{fig:example_part5}}\hspace{-8pt}\quad
         \centering
    \subfloat[Within the selected coarse cell, the fine cell with the highest utility is selected for tree repairing.]{
         \includegraphics[width=0.24\textwidth]{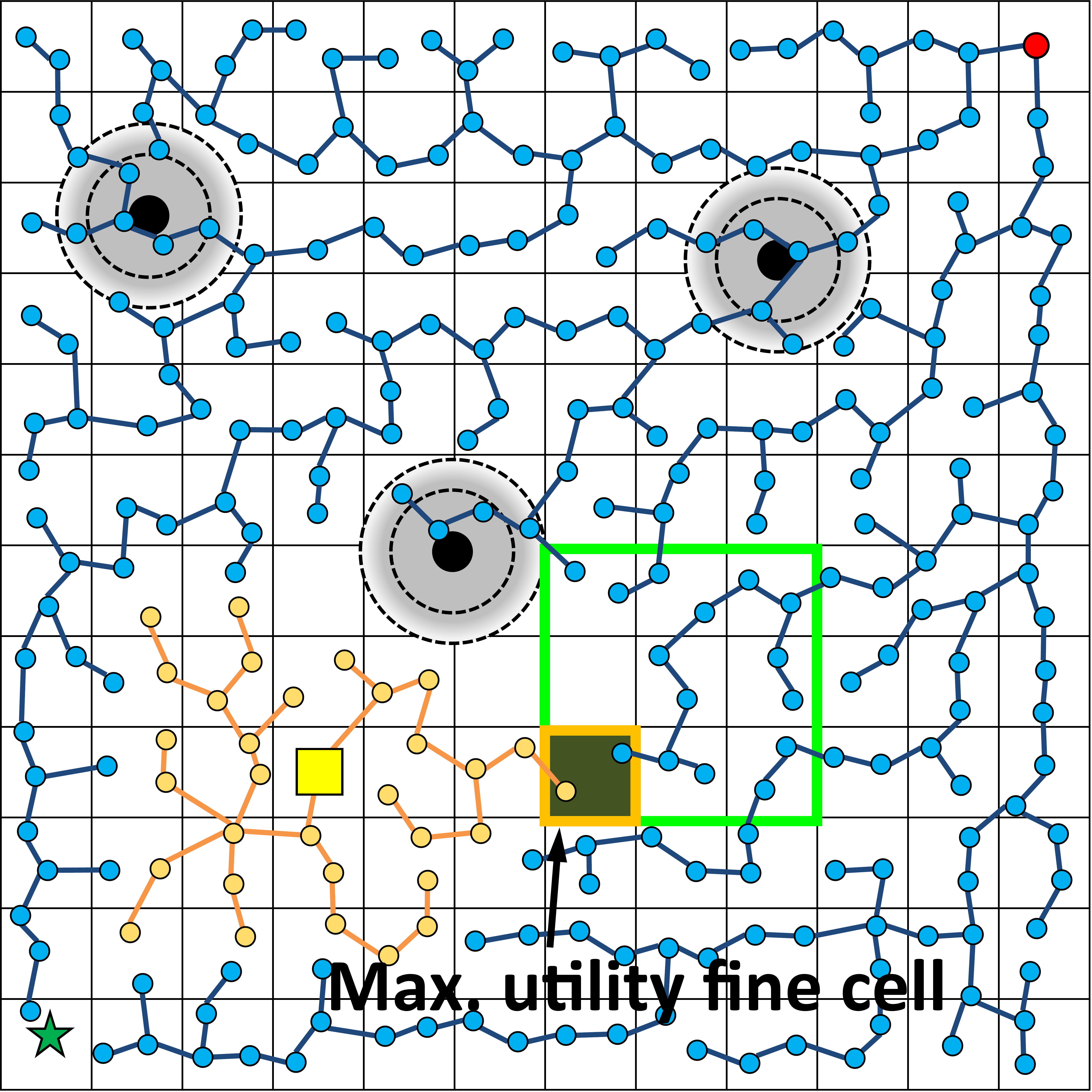}\label{fig:example_part6}}\hspace{-8pt}\quad
         \centering
    \subfloat[Tree is repaired by sampling and rewiring in the selected fine cell. Then, a new path is found.]{
         \includegraphics[width=0.24\textwidth]{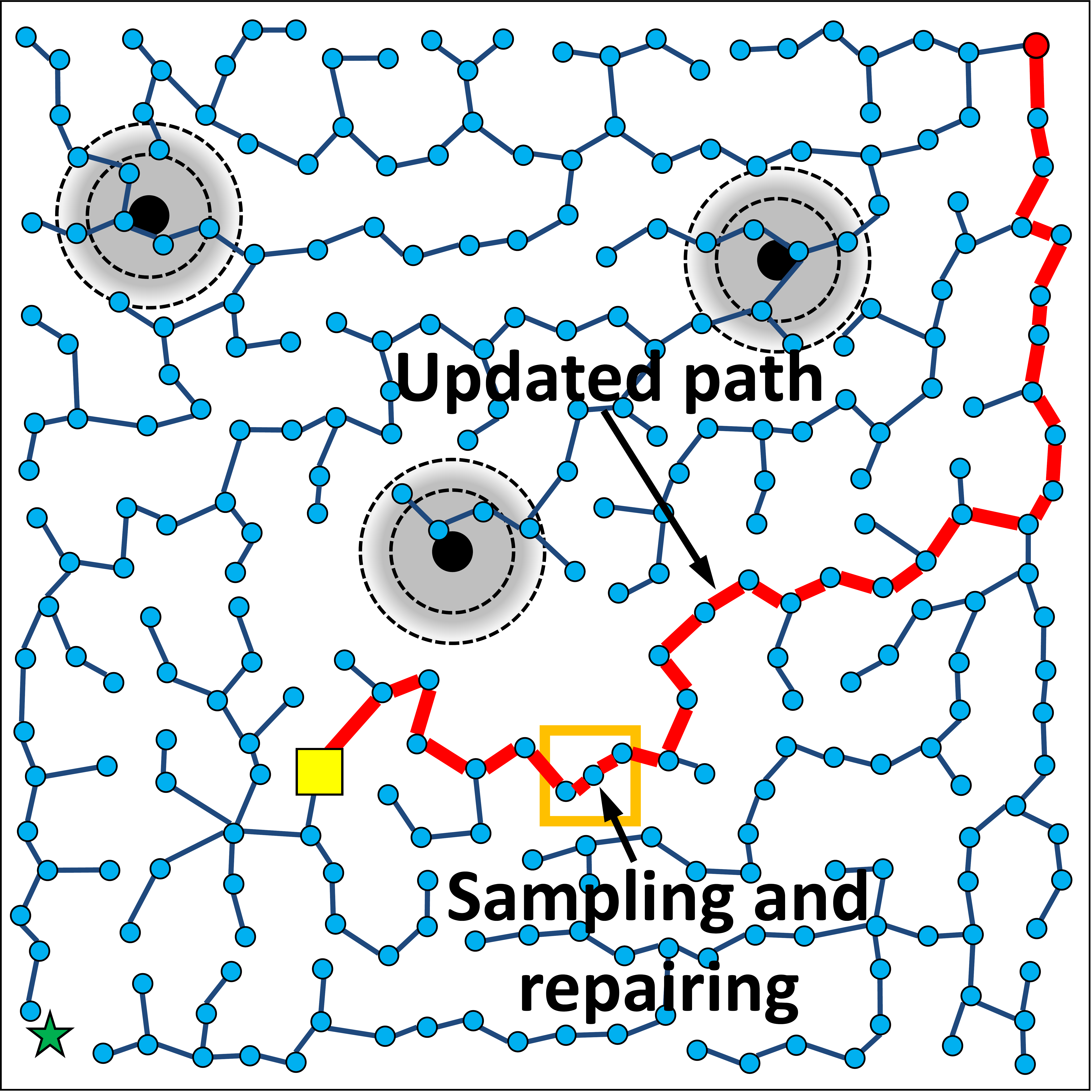}\label{fig:example_part7}}\hspace{-8pt}\quad
         \centering
    \subfloat[Tree is constantly rewired locally to find a better path to the goal upon reaching a path node.]{
        \includegraphics[width=0.24\textwidth]{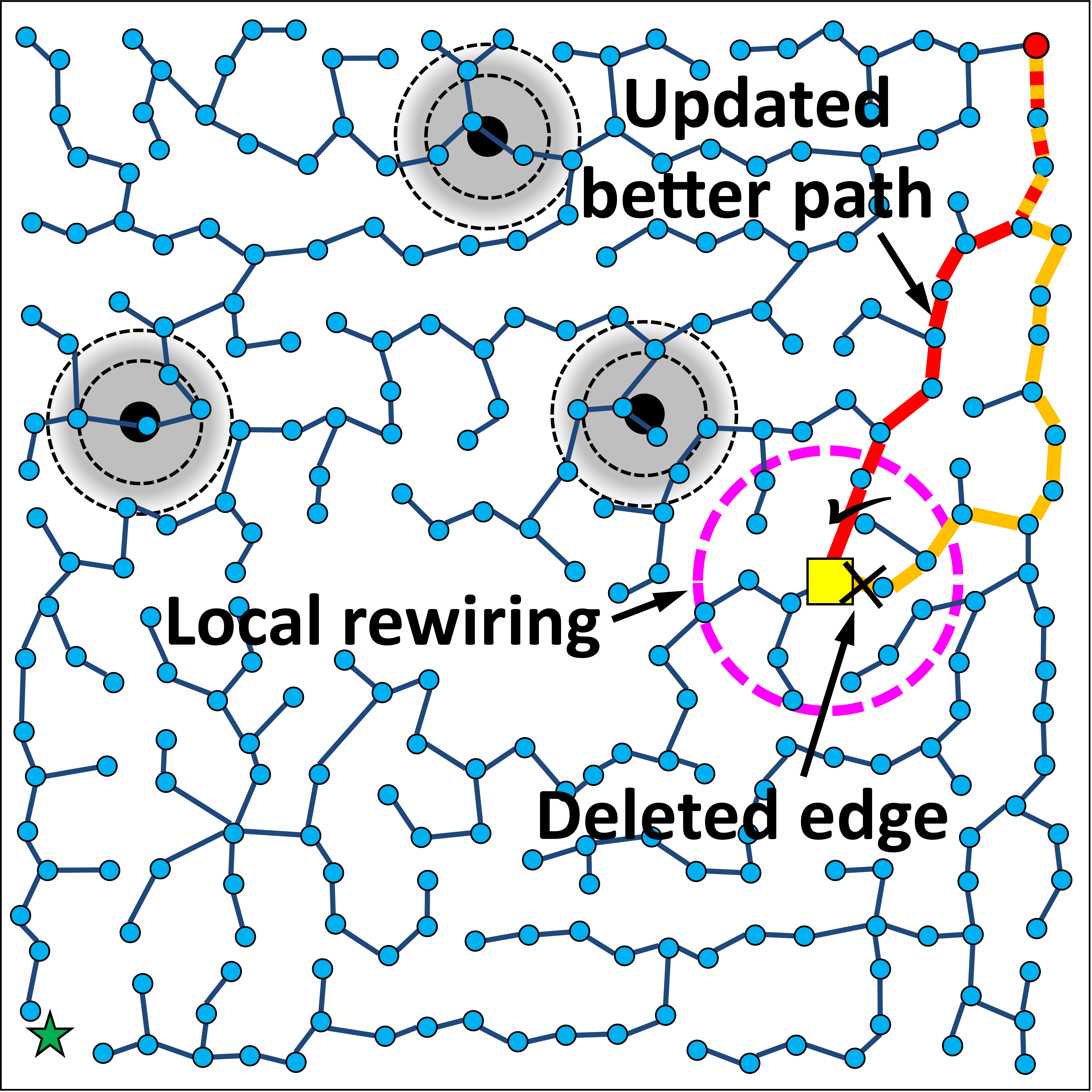}\label{fig:example_part8}}\\ 
        
          \caption{An illustrative example of SMARRT algorithm.}\label{fig:example}
 \end{figure*}

\subsubsection{Smart Sampling for Online Tree Repair}
\label{sampling_cell}
The multi-resolution utility map is then used  to find the best cell that 1) quickly repairs the tree and 2) guides the robot to the goal while avoiding the moving obstacle. Specifically, the method reads the utilities in the neighborhood of robot's current sitting cell at level $\ell=0$ and selects the cell with the most utility. If there is no valid cell nearby, the method switches to a coarser level to broaden the search incrementally until a valid cell is found. The selected cell for replanning is the finest cell with the most utility. Finally, a random sample is generated inside this cell to join to all the disjoint trees within its neighborhood. 

\subsection{Summary of SMARRT Algorithm}
Algorithm~\ref{alg:SMARRT} summarizes the SMARRT algorithm. Initially, the tree, $\mathcal{G}^1$, is rooted at the goal state $x_{goal}$, and explores the area for initial path denoted by $\sigma$ (\textbf{Line 1}). Then, the set of disjoint trees $\mathcal{G}$ is initialized as $\mathcal{G}^1$ (\textbf{Line 2}). At each iteration, obstacle position and robot position are updated (\textbf{Line 4} and \textbf{Line 5}). Partial path segment is checked against feasibility (\textbf{Line 6}). If the current path is feasible, the robot continues the movement along the path and attempts to search a better path nearby upon reaching a waypoint (\textbf{Line 7} and \textbf{Line 8}); otherwise, risky nodes are pruned from the tree and disjoint trees are identified by floodfill algorithm (\textbf{Line 10} and \textbf{Line 11}), meanwhile, multi-resolution utility map $\mathcal{U}$ is constructed (\textbf{Line 12}). The sampling cell is searched on the map in a bottom-up manner (\textbf{Line 13}). Finally, samples are generated within the sampling cell to repair the connectivity (\textbf{Line 14}).

\subsection{An Illustrative Example of SMARRT Algorithm}

Fig.~\ref{fig:example} illustrates an illustrative example of SMARRT algorithm in dynamic environment containing three moving obstacles. The robot is at the bottom left corner of the area. The feasibility checking horizon is represented by green circle. As shown in Fig.~\ref{fig:example}(a), the tree is rooted at the goal state and explores the search area by using RRT algorithm to find an initial path marked by red color. As seen in Fig.~\ref{fig:example}(b), the path segment close to robot is infeasible and the feasibility checking horizon is constructed to prune the risky nodes from the tree, resulting in two disjoint trees as shown in Fig.~\ref{fig:example}(c). Then the multi-resolution utility map with three levels is constructed and the selected sampling cell is searched hierarchically on the map, as seen in Fig.~\ref{fig:example}(d)-(f). After that, sample is created inside the sampling cell to repair connectivity between trees and a new feasible path is obtained, as shown in Fig.~\ref{fig:example}(g). Upon reaching a path node, the tree is rewired locally to find a better path to the goal as shown in Fig.~\ref{fig:example}(h).


         
 
 \begin{algorithm}[t] 
\caption{SMARRT Algorithm.} 
\label{alg:SMARRT} 
\begin{algorithmic}[1] 
\State $[\mathcal{G}^1,\sigma] \leftarrow$ \textbf{InitialPlan}($x_{robot},x_{goal},{\mathcal{X}}_{free},{\mathcal{X}}_{obs}$);
\State $\mathcal{G} \leftarrow \left \{ \mathcal{G}^1 \right \}$;
\While{$x_{robot} \neq x_{goal}$}
\State $\mathcal{X}_{obs} \leftarrow$ \textbf{UpdateObstacle}();
\State $\mathcal{X}_{robot} \leftarrow$ \textbf{UpdateRobot}();
\If {\textbf{isFeasible}($\sigma$) }
\State \textbf{MoveAlong}($\sigma$);
\State $\sigma \leftarrow$\textbf{BetterPathSearch}($\mathcal{G},\mathcal{X}_{obs},\sigma$);
\Else
\State $\mathcal{G} \leftarrow$\textbf{TreePruning}($\mathcal{G},\mathcal{X}_{obs},r_h$);
\State $\mathcal{G} \leftarrow$\textbf{DisjointTreeIdentify}($\mathcal{G}$);
\State $\mathcal{U}\leftarrow$ \textbf{UtilityMapConstruct}($\mathcal{G}$);
\State $\tau^{s}\leftarrow$\textbf{SamplingCellSearch}($\mathcal{U}$);
\State $\mathcal{G},\sigma \leftarrow$\textbf{TreeRepairing}($\mathcal{G},\mathcal{X}_{obs},\tau^{s}$);
\EndIf
\EndWhile
\end{algorithmic}
\end{algorithm}

        

\begin{figure*}

\centering
\begin{subfigure}{\textwidth}
  \begin{minipage}{\textwidth}\footnotesize
  \centering
    \subsubfloat{\includegraphics[width=0.5\textwidth]{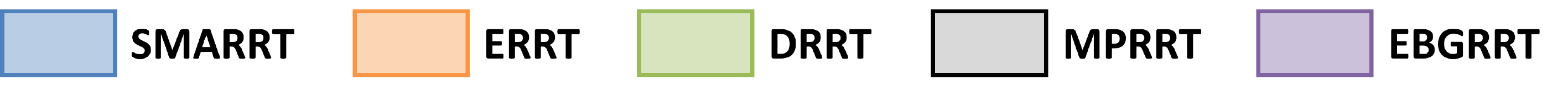}}{}\\
  \subsubfloat{\includegraphics[width=0.32\textwidth]{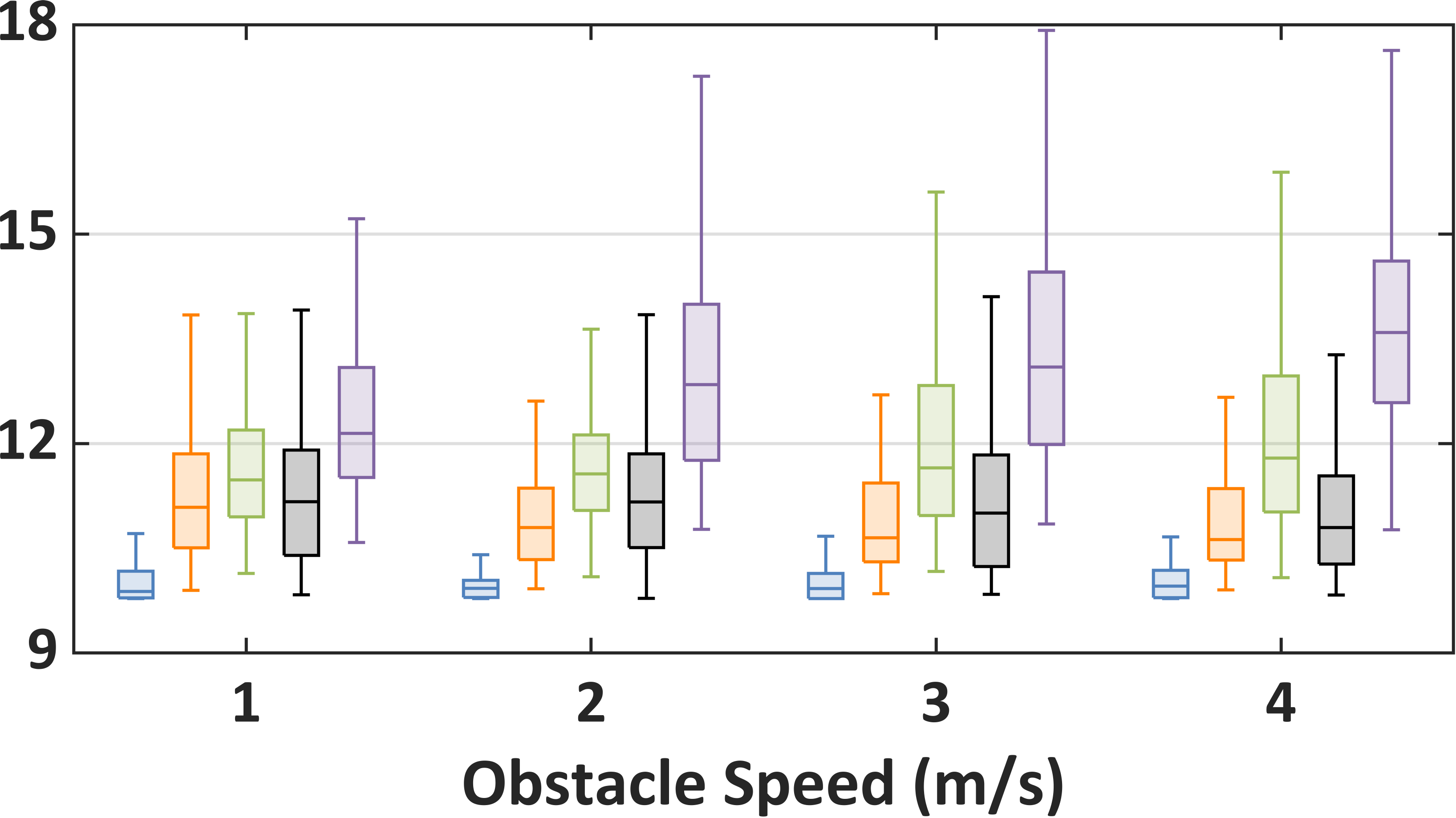}}{Travel time (s)}\hspace{-10pt}
  \qquad
  \subsubfloat{\includegraphics[width=0.32\textwidth]{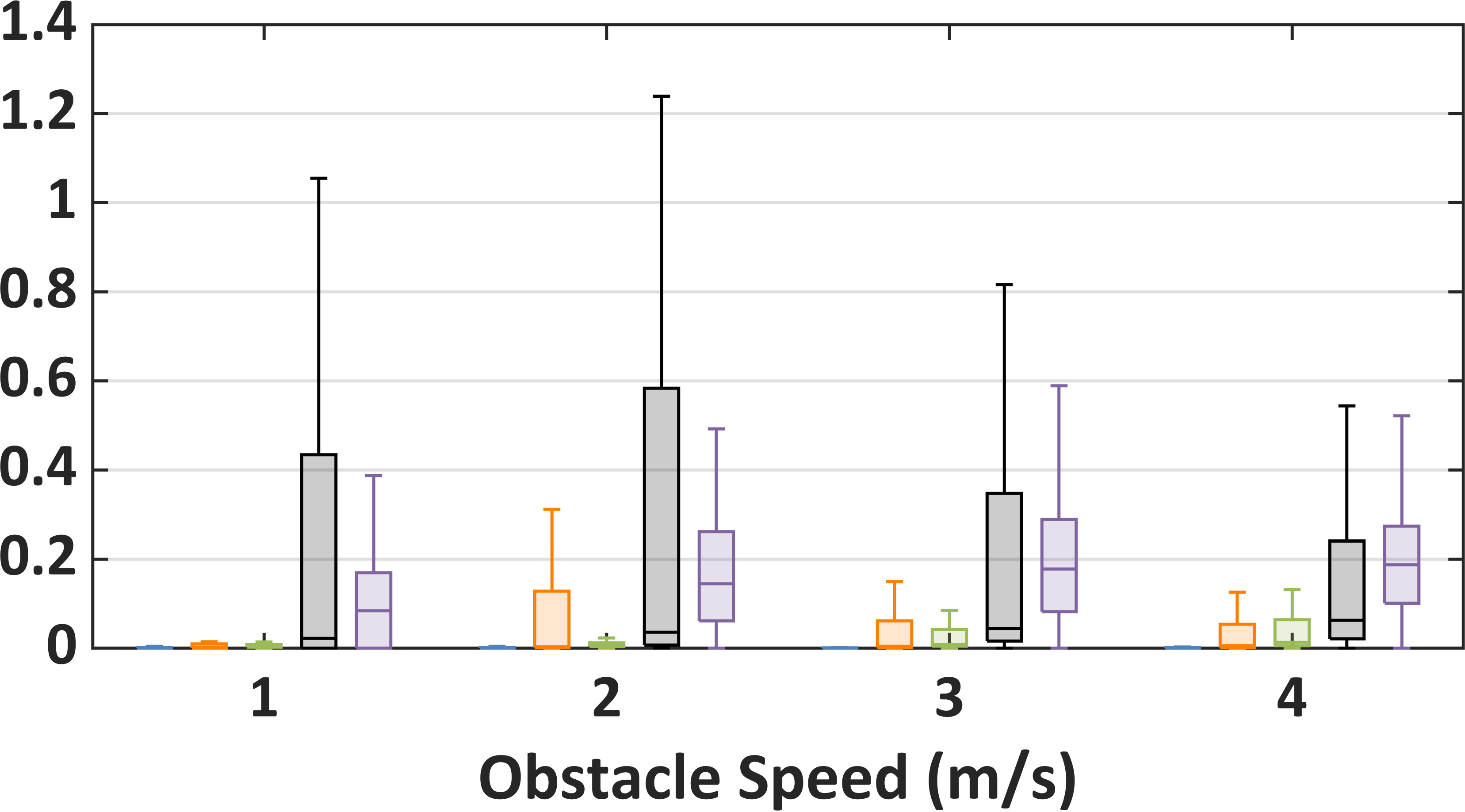}}{Average replanning time (s)}\hspace{-10pt}
  \qquad
  \subsubfloat{\includegraphics[width=0.32\textwidth]{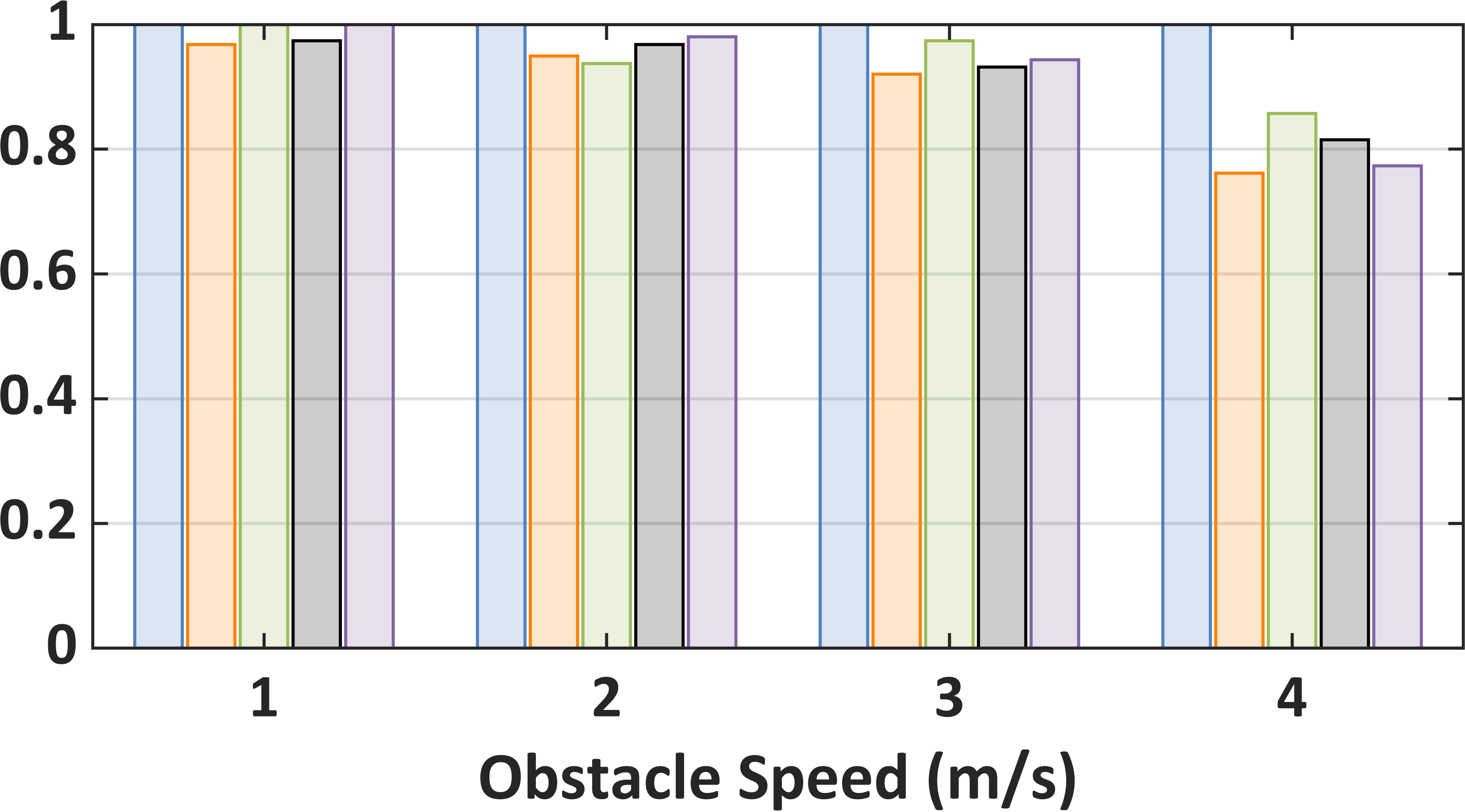}}{Success rate}
  \end{minipage}
\caption{Monte Carlo Simulation with three moving obstacles}\label{mc_part1}
\end{subfigure}

\centering
\begin{subfigure}{\textwidth}
  \begin{minipage}{\textwidth}\footnotesize
  \centering
  \subsubfloat{\includegraphics[width=0.32\textwidth]{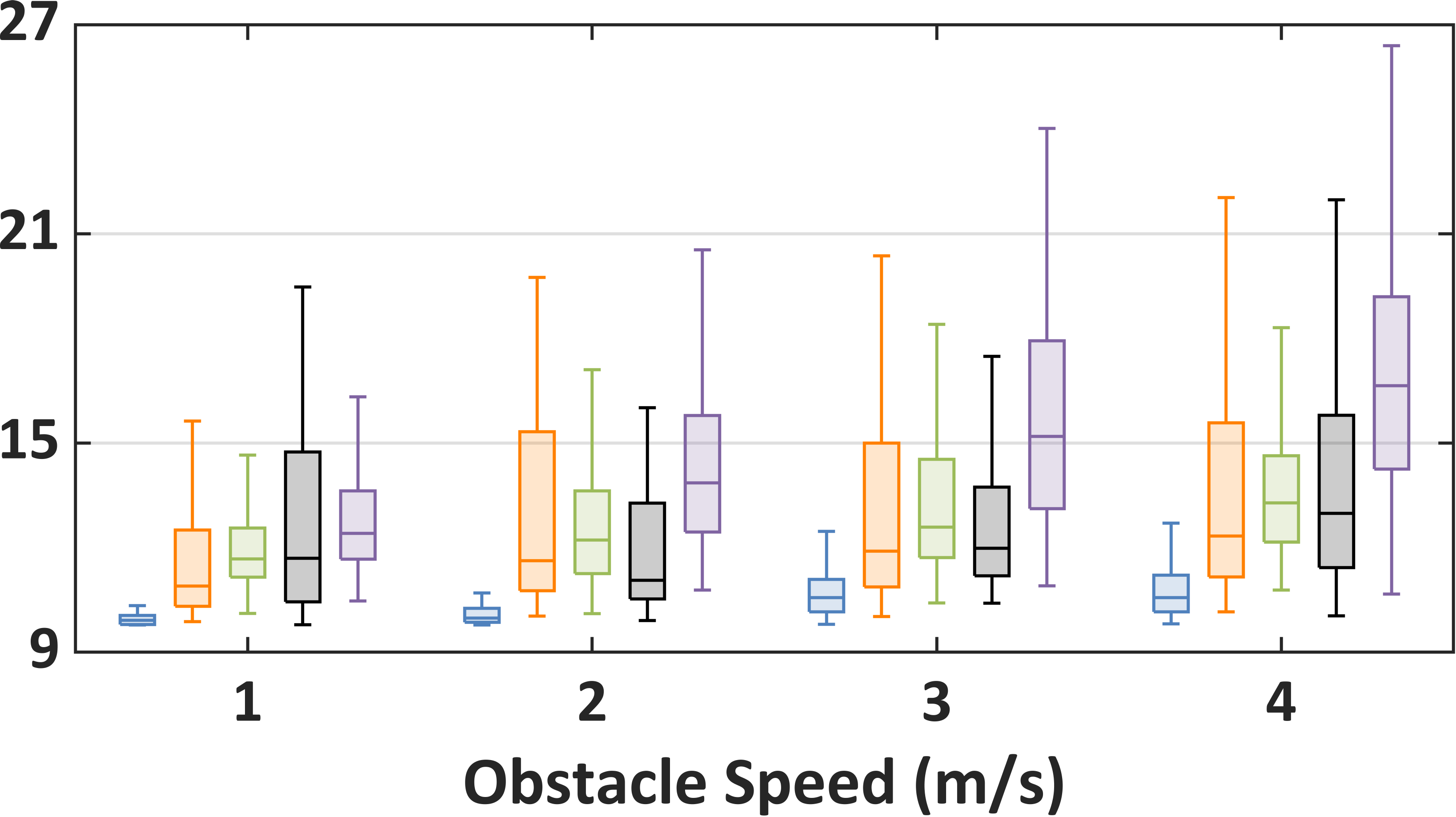}}{Travel time (s)}\hspace{-10pt}
  \qquad
  \subsubfloat{\includegraphics[width=0.32\textwidth]{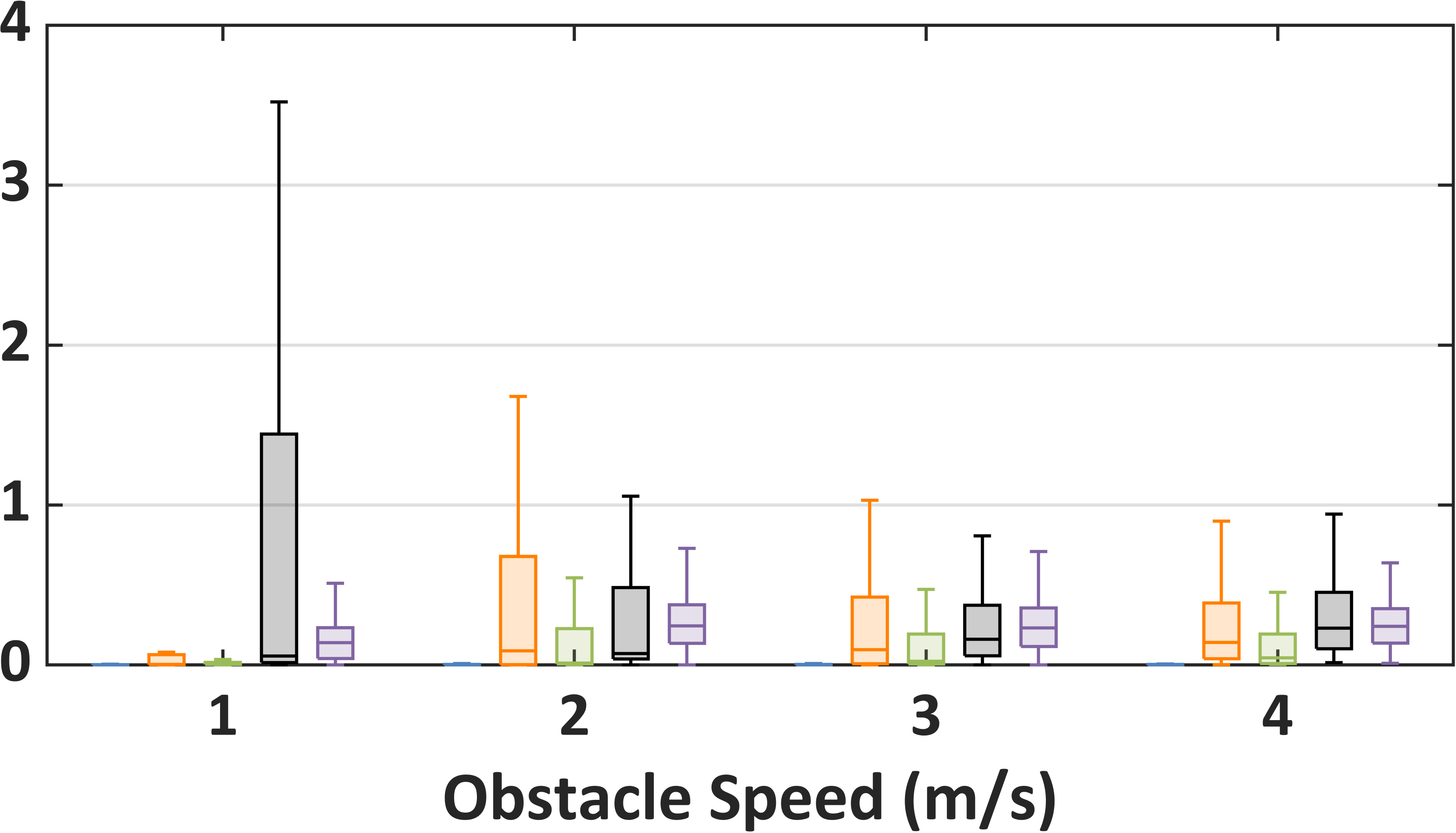}}{Average replanning time (s)}\hspace{-10pt}
  \qquad
  \subsubfloat{\includegraphics[width=0.32\textwidth]{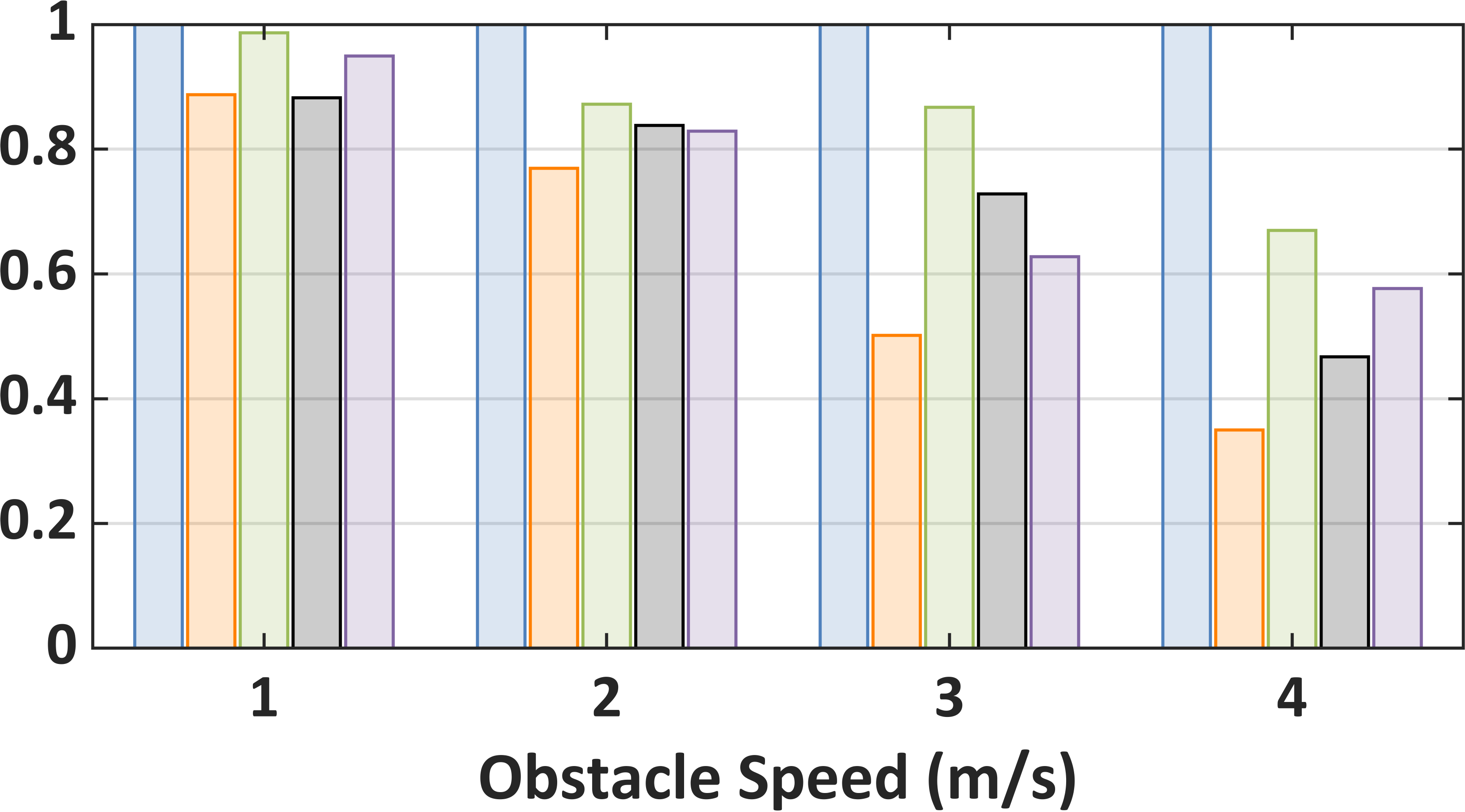}}{Success rate}
  \end{minipage}
\caption{Monte Carlo Simulation with six moving obstacles}\label{mc_part2}
\end{subfigure}

\centering
\begin{subfigure}{\textwidth}
  \begin{minipage}{\textwidth}\footnotesize
  \centering
  \subsubfloat{\includegraphics[width=0.32\textwidth]{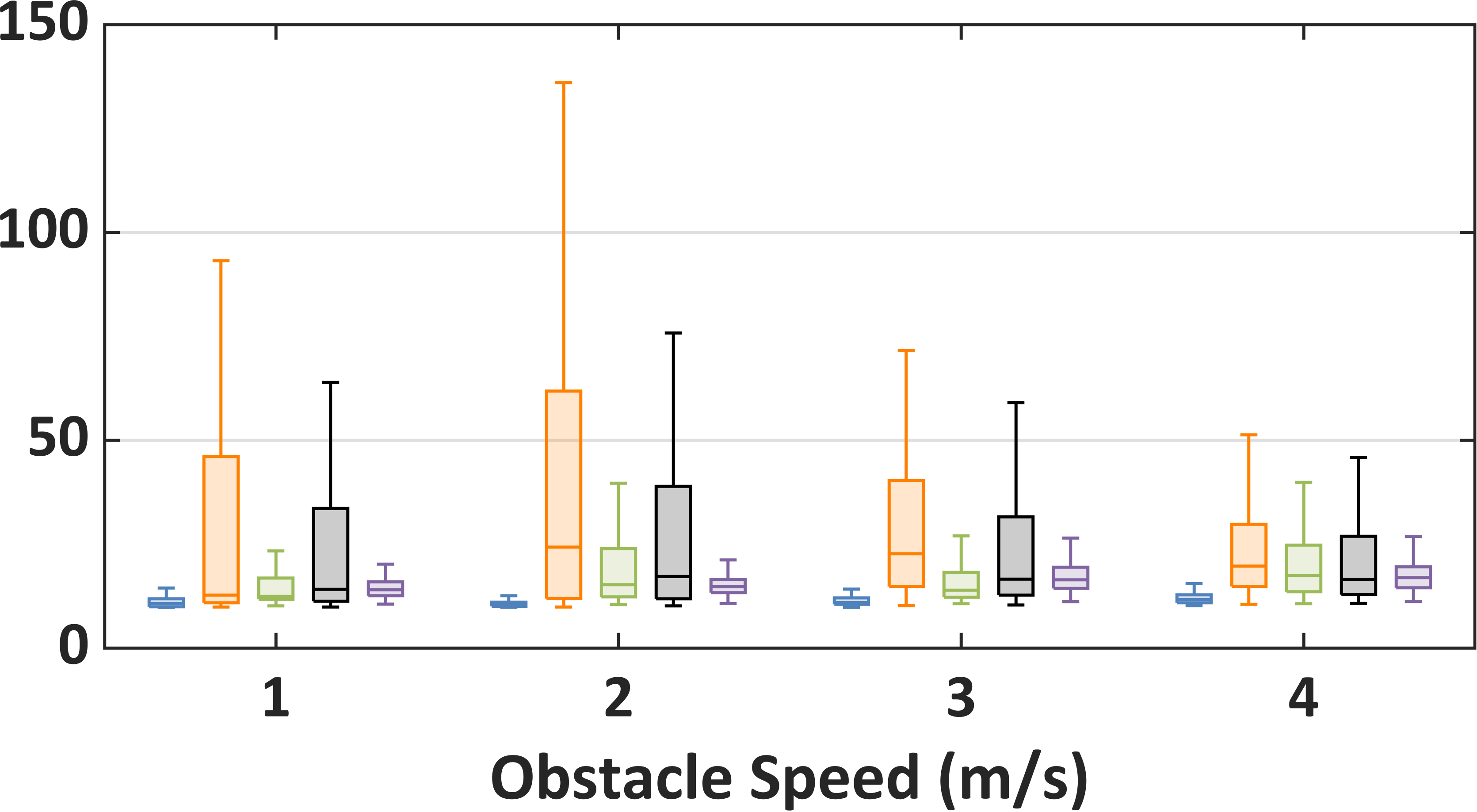}}{Travel time (s)}\hspace{-10pt}
  \qquad
  \subsubfloat{\includegraphics[width=0.32\textwidth]{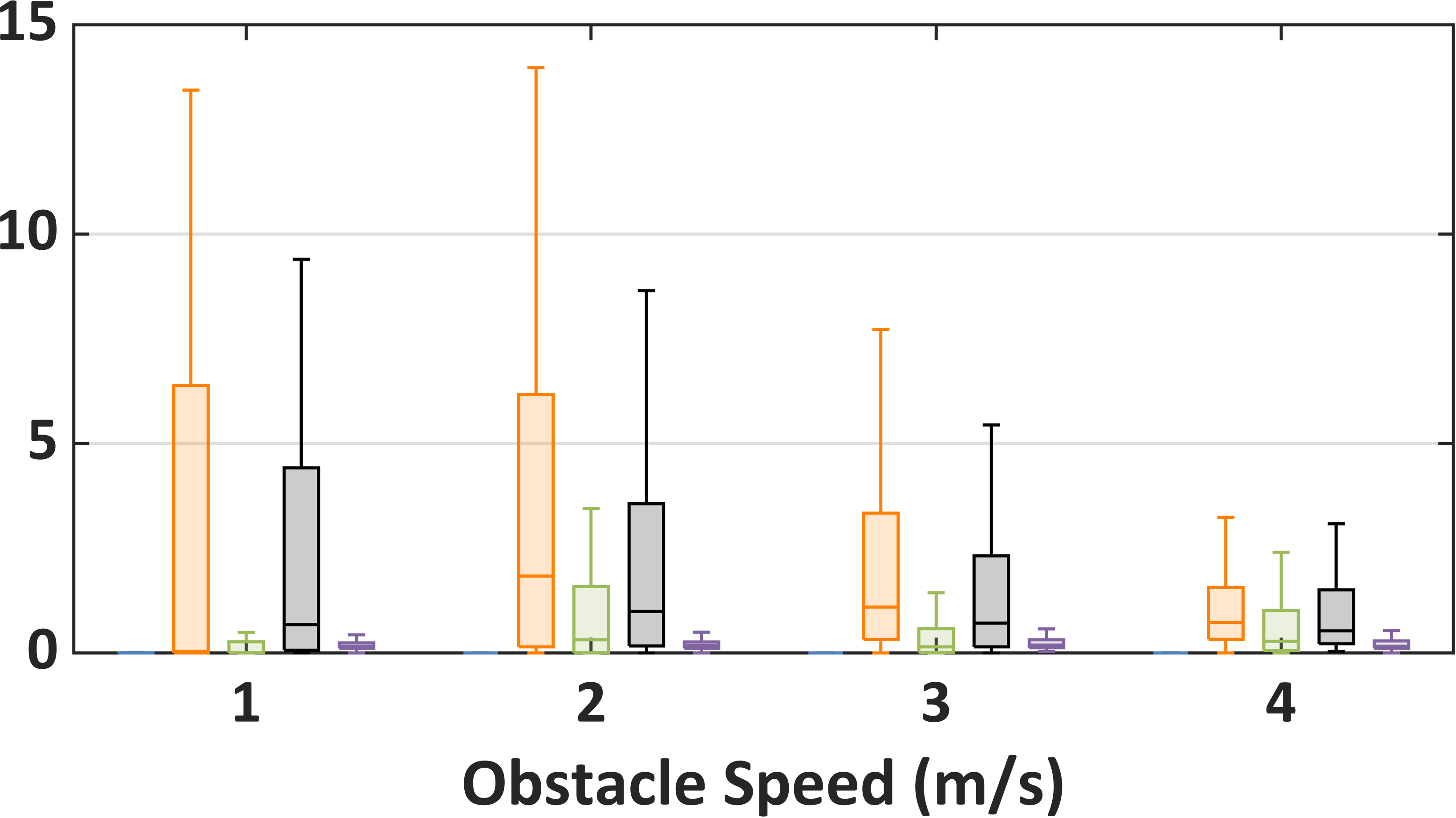}}{Average replanning time (s)}\hspace{-10pt}
  \qquad
  \subsubfloat{\includegraphics[width=0.32\textwidth]{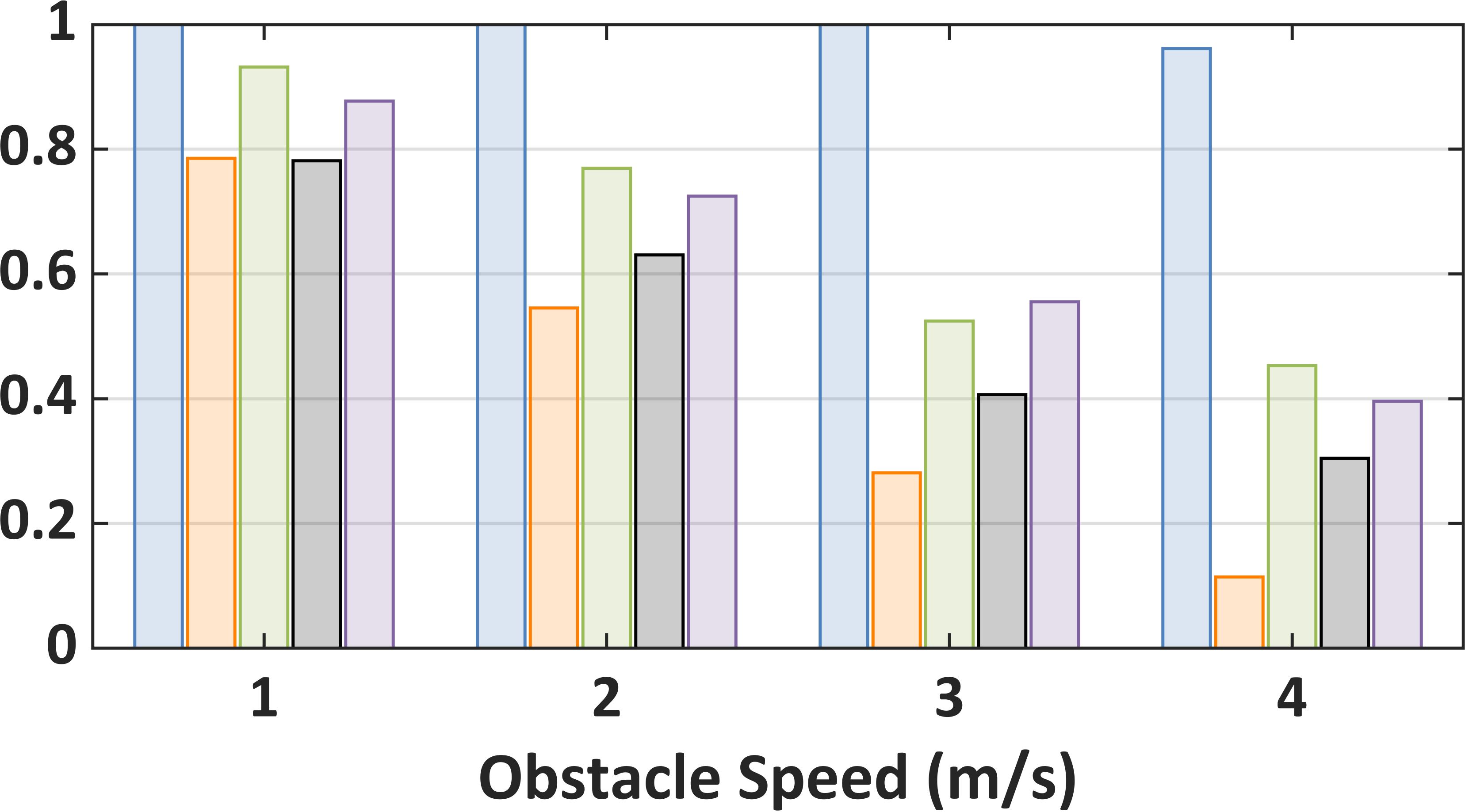}}{Success rate}
  \end{minipage}
\caption{Monte Carlo Simulation with nine moving obstacles}\label{mc_part3}
\end{subfigure}

\caption{Monte Carlo Simulation Results.}\label{mc}
\end{figure*}

\section{Results and Discussion}
\label{sec:results}

\subsection{Monte Carlo Analysis}

\subsubsection{Simulation Setup}

All simulations were done in C++ on Ubuntu with an Intel CPU. A robot was simulated as a mass point model with constant speed of $4m/s$. It is assumed that the robot knows the present locations of the moving obstacles. The initial position of robot is $(2, 30)m$ and the goal position is $(30, 2)m$. The steering step size for tree expansion is set as $2m$. Obstacles are circular with radius $1m$, and they move along a randomly selected heading on the interval $\left [0,2\pi  \right ]$ for a random distance up to $10m$. If the obstacles touch the boundary, a new heading is randomly selected. The scenario is an open area of size $32m \times 32m$ and partitioned into a $32 \times 32$ tiling structure consisting of $1m \times 1m$ cells. This resulted in an MST with $L = 5$. Three, six, or nine moving obstacles were present on the map, and moved at a constant speed selected from $\left \{1,2,3,4 \right\}m/s$, resulting in a total of $12$ combinations. For each combination of obstacle number and speed, 5 random scenarios of the moving obstacles are generated and fixed. Then, 30 Monte-Carlo simulations were performed on each scenario for a total number of $150$ trials per obstacle number and speed combination. If the robot collides with the obstacles, the trial ends in failure. 

\subsubsection{Performance Metrics}

The SMARRT algorithm is evaluated based on the following performance metrics:

\begin{itemize}
\item Travel time: The total time it takes for the robot to travel from the start position to the goal.

\item Average replanning time: the average time spent replanning a new path if the existing path becomes infeasible.

\item Success rate: The number of times the robot successful collision-free runs divided by the total number of runs.

\end{itemize}

\subsubsection{Competing Reactive Path Planners}

The performance of the SMARRT algorithm is compared against four other feasible sampling-based replanning algorithms: Extended RRT (ERRT)\cite{bruce2002real}, Dynamic RRT (DRRT)\cite{ferguson2006replanning}, Multipartite RRT (MP-RRT)\cite{zucker2007multipartite}, and Efficient Bias-goal Factor RRT (EBG-RRT)\cite{yuan2020efficient}.

\subsubsection{Monte Carlo Simulation Results} The results are shown in Fig.~\ref{mc}, where the SMARRT algorithm outperforms other alternative methods in terms of travel time, average replanning time, and success rate. Specifically, SMARRT algorithm achieve $100\%$ success rate in most scenarios and high success rate in the worst scenario since it can react to the obstacle and replan a new feasible path quickly. The median value of average replanning time is shown in Table~\ref{tab:time}. Clearly, SMARRT requires the least amount of time to replan and avoid the obstacles, contributing to its very high success rate. Moreover, the travel time of SMARRT algorithm is lower than other methods since the SMARRT algorithm creates sample in the cell with highest utility and it rewires the tree to find a better path to the goal constantly.

\begin{table}[ht!]
\caption{Median of Average Replanning Time (s)}
\label{tab:time}
\begin{tabular}{llllll}
\hline
Scenario & SMARRT & ERRT & DRRT & \tabincell{l}{MP-\\RRT} & \tabincell{l}{EBG-\\RRT} \\ \hline
3 obstacles, 1m/s & 0.00035 & 0.0025 & 0.0057 & 0.0219 & 0.0841 \\ \hline
3 obstacles, 2m/s & 0.00038 & 0.0035 & 0.0067 & 0.0362 & 0.1447 \\ \hline
3 obstacles, 3m/s & 0.00032 & 0.0043 & 0.0084 & 0.0445 & 0.178 \\ \hline
3 obstacles, 4m/s & 0.00038 & 0.0056 & 0.0131 & 0.0627 & 0.1875 \\ \hline

6 obstacles, 1m/s & 0.00055 & 0.0033 & 0.0068 & 0.0562 & 0.1398 \\ \hline
6 obstacles, 2m/s & 0.0012 & 0.0892 & 0.0124 & 0.0721 & 0.2444 \\ \hline
6 obstacles, 3m/s & 0.0019 & 0.0961 & 0.0233 & 0.1609 & 0.2318 \\ \hline
6 obstacles, 4m/s & 0.002 & 0.1416 & 0.0444 & 0.231 & 0.2413 \\ \hline

9 obstacles, 1m/s & 0.0044 & 0.0344 & 0.008 & 0.6786 & 0.1648 \\ \hline
9 obstacles, 2m/s & 0.0034 & 1.836 & 0.3143 & 0.9904 & 0.1856 \\ \hline
9 obstacles, 3m/s & 0.0035 & 1.098 & 0.1483 & 0.7163 & 0.191 \\ \hline
9 obstacles, 4m/s & 0.0035 & 0.7288 & 0.2767 & 0.5274 & 0.1603 \\ \hline
\end{tabular}
\end{table}

\subsection{Application to Smart Factory}

The Smart Factory presents a clear use case for efficient replanning algorithms.
Originally proposed in the mid-2000's, the Smart Factory platform is intended as as way to integrate the Internet of Things into manufacturing via adaptable and context-sensitive resource management\cite{lucke2008d}.
Developments in Internet of Things (IoT) and Cloud Computing technology platforms have allowed for an expanded academic focus on the Smart Factory in a variety of fields\cite{rub2019j}. 
Common features of Smart Factory models include\cite{wang2016s}:
\begin{itemize}
    \item Adaptive and diverse resource management.
    \item Dynamic routing of the manufacturing process to ensure efficiency and robustness.
    \item Total network integration between machines, products, workers, and information systems.
    \item Data informed Self-Organization of manufacturing systems.
\end{itemize}

\begin{figure*}[ht!]
    \centering
    \subfloat[Initial path is obtained]{
    \includegraphics[width = 0.32\textwidth]{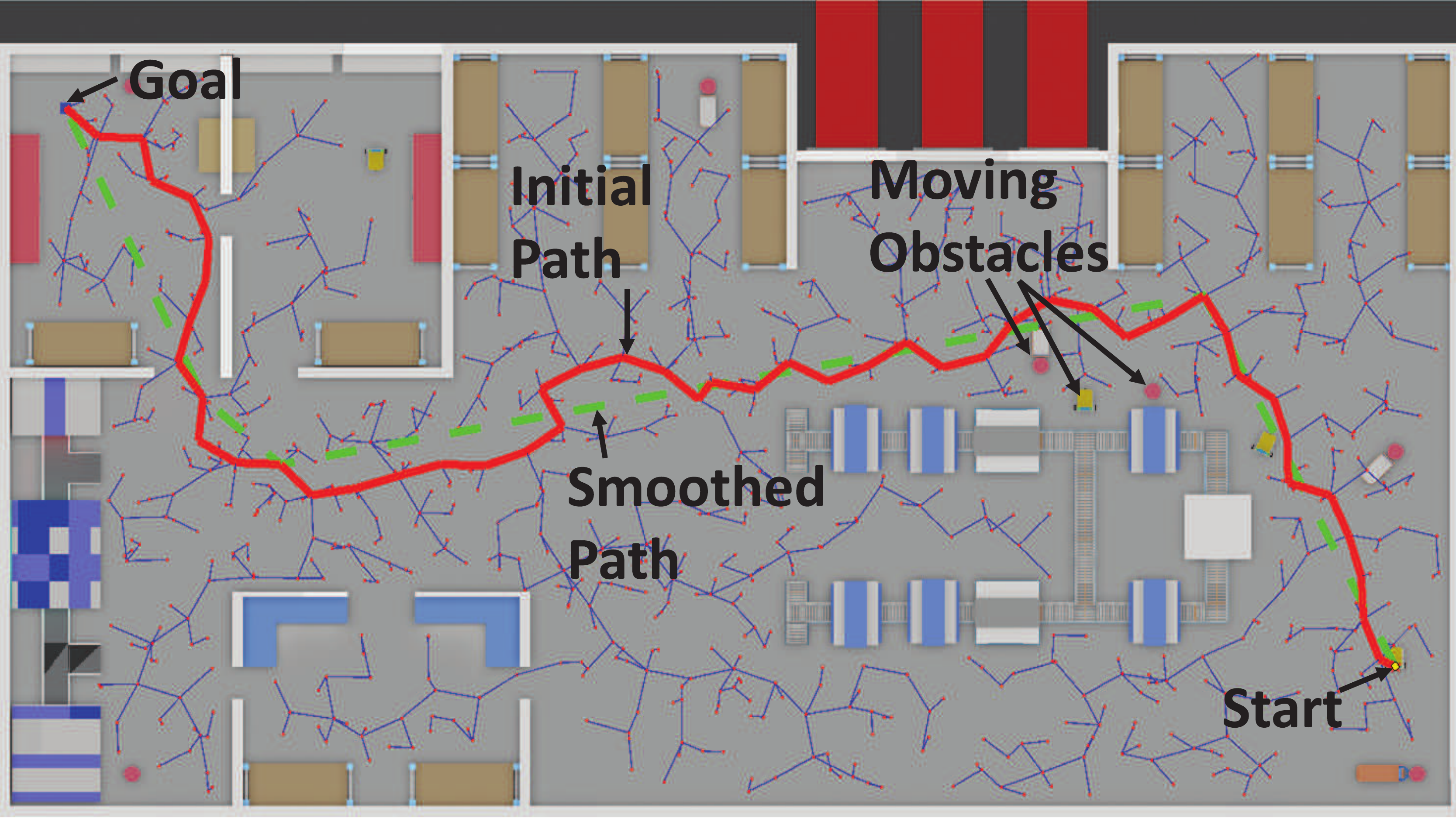}}
    \subfloat[Path is broken via Obstacle Collision Zone]{
    \includegraphics[width = 0.32\textwidth]{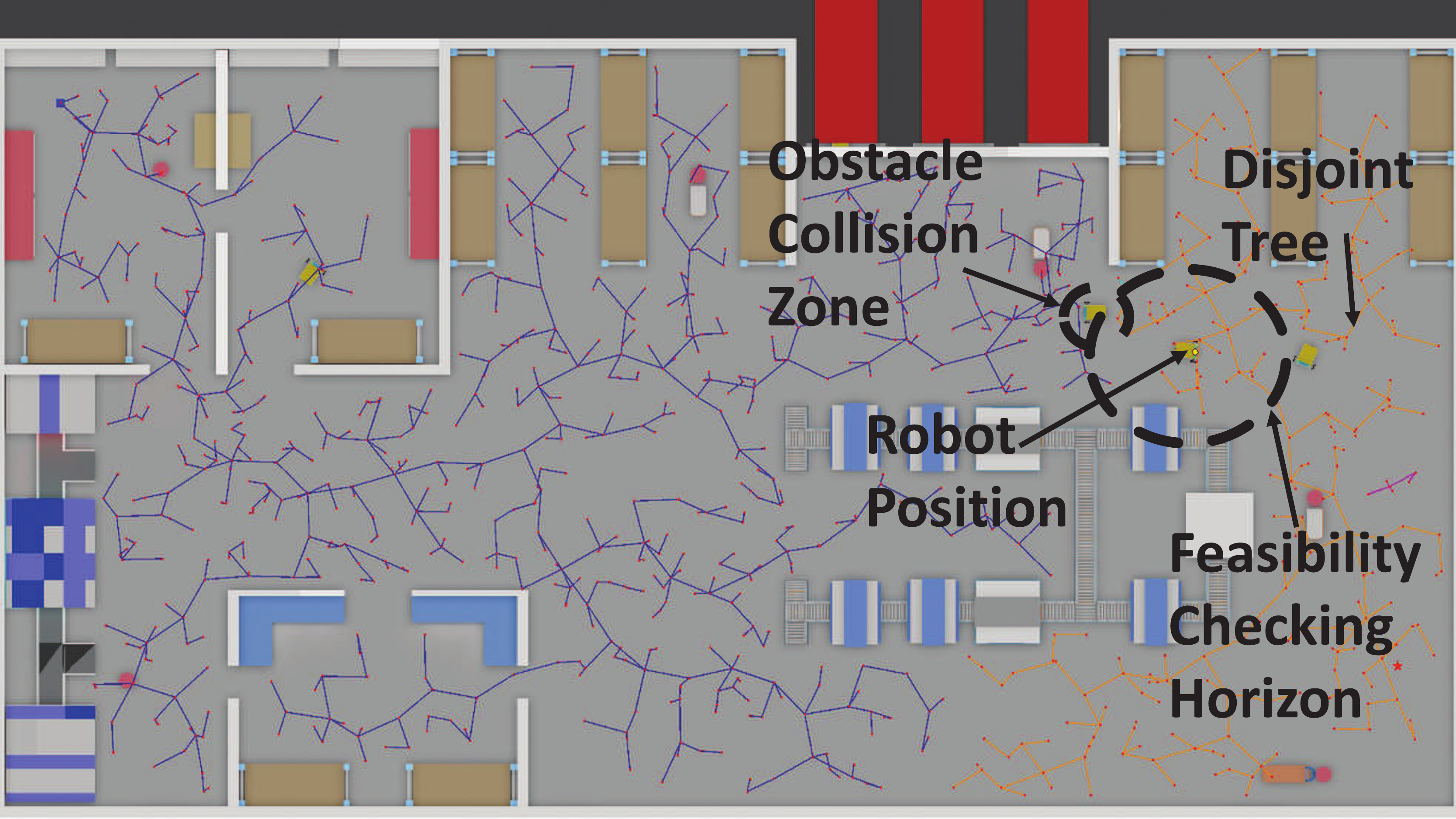}}
    \subfloat[Path is repaired via sample and edge creation to connect the two disjoint trees]{
    \includegraphics[width = 0.32\textwidth]{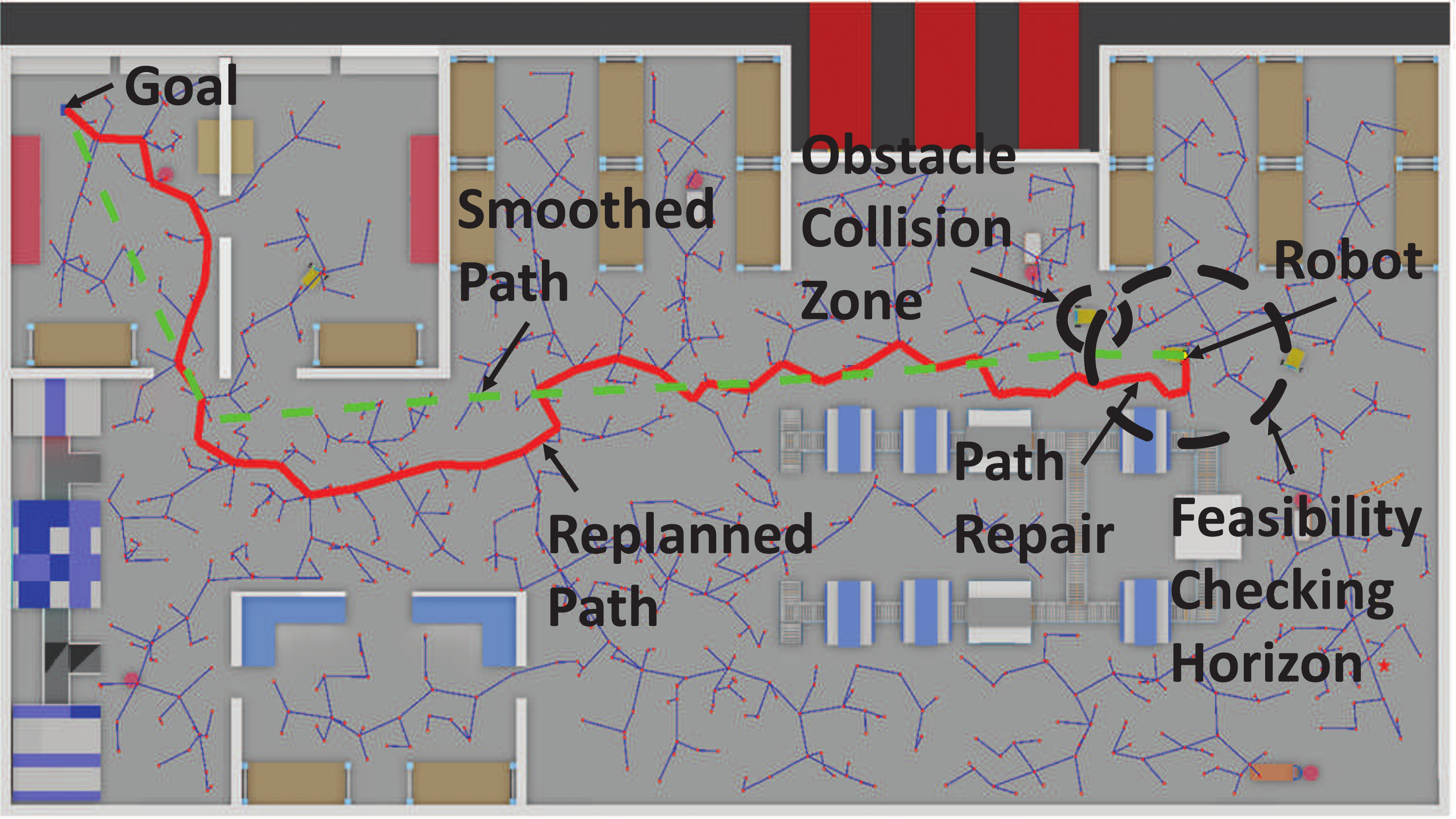}}
    \caption{Demonstration of SMARRT within Simulated Smart Factory environment}
    \label{fig:SmartFactory}
\end{figure*}

The requirement for dynamic routing of goods provides a strong framework for implementation of Autonomously Guided Vehicle (AGV) path planning as spontaneous changes to resource needs occur.
AGV navigation in the Smart Factory environment requires the ability to form multi-step paths while avoiding both static and dynamic obstacles. 
Given that safety is paramount, regular feasibility checking and efficient replanning are necessary to both ensure collision avoidance while also reducing unnecessary travel time.

A Smart Factory simulation environment was developed as a simulated space to apply SMARRT and examine its qualitative performance within a setting where we could expect an efficient path planner to be reasonably applied.
The space (Fig ~\ref{fig:SmartFactory}) corresponds to a $66m \times 38m$ manufacturing environment. 
Prominent features of the space include two networks of conveyor fed Smart Machines (one central within the space and one against the far left wall), two distinct areas for storage of resources at various stages in the manufacturing process, and three distinct rooms for employees to process products as they are being produced.

The top-left corner of the space features two rooms which are designated as ``painting rooms" and contain models similar to automatic-spray painting machines and large industrial curing ovens.
There is a room designated for Quality Control and receiving of resources from the loading docks positioned in the top center of the space. 
The aisles between the storage shelves act as narrow passageways, a common failure point for reactive path planning algorithms.
Similar to real world manufacturing environments, there are also multiple possible points of entrance to the distinct rooms; allowing for a secondary option for navigation provided the optimal entryway is temporarily blocked.
Overall, the space provides both a complex static geometry within which the simulated robot needs to navigate, and a coherent application within which the SMARRT algorithm can be tested against various real-world challenges.

Beyond the static obstacle avoidance, there are a variety of dynamic obstacles which must be avoided during the navigation process. The models considered include representations of human workers, both at stations and moving throughout the space, as well as other robots on paths unknown to the controlled AGV. 

As seen in Fig.~\ref{fig:SmartFactory}, the SMARRT algorithm is able to perform the required task and reach its destination despite the constrained environment and the many moving obstacles. First, the initial tree is constructed and the robot begins moving to the goal. The robot only replans the path when nearby moving obstacles obstruct its path, and the robot is able to successfully replan its path and reach the goal.
Including the time necessary to sample the space, the entire task is completed in the equivalent of 28.3 seconds of real time operation.
Replanning occurs rapidly, with the displayed path repair (Fig ~\ref{fig:SmartFactory} c) taking a total of 0.01 seconds.

\section{Conclusions and Future Work}
\label{sec:conclusions}

This paper presents a sampling-based path replanning algorithm, called SMARRT, for fast replanning in dynamic environments. SMARRT checks the feasibility of the path and tree online within the nearby vicinity of the robot, prunes the infeasible parts that have a high risk of colliding with nearby obstacles, and constructs a multi-resolution utility map for fast replanning. It is shown that SMARRT is computationally efficient and provides a very high success rate of reaching the destination without collisions to the obstacles. The algorithm is comparatively evaluated with four existing algorithms via Monte Carlo Simulation in complex dynamic environments. The results show that SMARRT results in reduced travel times, replanning times, and higher success rates.


 \balance
\bibliographystyle{ieeetr}
\bibliography{reference}
\end{document}